\pdfoutput=1

\documentclass[11pt]{article}

\usepackage[final]{acl}

\usepackage{times}
\usepackage{latexsym}
\usepackage[T1]{fontenc}
\usepackage{booktabs} 

\usepackage[utf8]{inputenc}

\usepackage{microtype}
\usepackage{tcolorbox}

\usepackage{inconsolata}
\usepackage{hyperref}
\usepackage{graphicx}
\usepackage{tikz}

\usepackage{multirow}
\usepackage{xcolor}
\usepackage{tablefootnote}
\usepackage{placeins}
\usepackage{listings}
\definecolor{Green}{RGB}{0,255,0}
\definecolor{YellowGreen}{RGB}{34,139,34}
\definecolor{Red}{RGB}{255, 0, 0}
\definecolor{Grey}{RGB}{128, 128, 128}
\title{Instructions for *ACL Proceedings}

\author{Argyrios Papoudakis \qquad Mirella Lapata \qquad Frank Keller \\
Institute of Language, Cognition and Computation \\ 
School of Informatics, University of Edinburgh \\
10 Crichton Street, Edinburgh EH8 9AB \\
\texttt{a.papoudakis@sms.ed.ac.uk}, \texttt{\{mlap, keller\}@inf.ed.ac.uk}}
\title{\textsc{BookWorm}: A Dataset for Character Description and Analysis}

\begin{document}
\maketitle
\begin{abstract}

Characters are at the heart of every story, driving the plot and engaging readers. In this study, we explore the understanding of characters in full-length books, which contain complex narratives and numerous interacting characters. We define two tasks: \emph{character description}, which generates a brief factual profile, and \emph{character analysis}, which offers an in-depth interpretation, including character development, personality, and social context. We introduce the \textsc{BookWorm} dataset, pairing books from the Gutenberg Project with human-written descriptions and analyses. Using this dataset, we evaluate state-of-the-art long-context models in zero-shot and fine-tuning settings, utilizing both retrieval-based and hierarchical processing for book-length inputs. Our findings show that retrieval-based approaches outperform hierarchical ones in both tasks. Additionally, fine-tuned models using coreference-based retrieval produce the most factual descriptions, as measured by fact- and entailment-based metrics. We hope our dataset, experiments, and analysis will inspire further research in character-based narrative understanding.

\end{abstract}

\begin{figure*}[t]
    \centering
    \footnotesize
    \begin{tabular}{@{}p{16cm}@{}} \toprule
    \multicolumn{1}{c}{\textbf{\textit{Book}:} Bleak House by Charles Dickens,  \textbf{\textit{Character}:} Esther Summerson} \\ \hline
    \textbf{\textit{Description}:}
     The narrator and protagonist. Esther, an orphan, becomes the housekeeper at Bleak House when she, Ada, and Richard are taken in by Mr. Jarndyce. Everyone loves Esther, who is \textcolor{YellowGreen}{selfless} and \textcolor{YellowGreen}{nurturing}, and she becomes the confidante of several young women. Although she eventually does find her mother, circumstances prevent them from developing a relationship. \textcolor{Red} {At first a hesitant, insecure narrator, Esther’s confidence in her storytelling grows}, and she controls the narrative skillfully. \\ \hline
    \textbf{\textit{Analysis}:} 
    Esther Summerson, the narrator and protagonist of Bleak House, [..] she proves to be a confident narrator who never misses the opportunity to relate others’ compliments of her.[..] \textcolor{Red} {As her narrative gains breadth and depth, her confidence as a narrator grows. She deliberately withholds information or delays including it to give her story coherence and dramatic effect.} And even though she is for the most part a reliable narrator (a narrator we can trust to accurately tell the story), she is less reliable when relaying information about her romantic life. Esther nurtures everyone around her, and her first instinct is to \textcolor{YellowGreen}{be motherly, perhaps because she has never had a caring mother figure of her own.} [..] Ironically, Esther, for all her caring and tenderness, is the unwitting cause of great unhappiness. [..] Because of Esther’s illegitimate birth, Lady Dedlock was forever estranged from her sister, Miss Barbary, and was forced to carry a painful secret. Because other unhappinesses, [..] \textcolor{Grey}{we could argue that Esther is indirectly responsible for these as well.} \\ \bottomrule
    \end{tabular}
    \caption{Examples of character description and analysis. Both refer to the transformation of Esther Summerson from a hesitant to a confident narrator. However, the analysis provides more detail focusing on her skill as a narrator (red). The description includes Esther's attributes and behaviour, referring to her as a selfless and nurturing figure, while the analysis provides an interpretation of this trait based on her background (green). The character description briefly touches on Esther’s background, while the analysis demonstrates how, ironically and indirectly, she causes pain to others (grey), adding a moral and psychological dimension.}
    \label{fig:description_vs_analysis}
\end{figure*}

\section{Introduction}

Stories play a key role in shaping our understanding of the world, serving as a medium to share experiences, communicate ideas, teach, and entertain. The two main building blocks of every story are the plot and characters~\cite{phelan1989reading}. Characters are particularly important, as they form the primary means through which readers engage and relate to the story. 

Understanding characters is also necessary from a computational perspective, if models are to summarize, analyse, or generate stories effectively. Over the past decade, the field of natural language processing has developed computational methods to understand narratives from a character-centric perspective. Previous work has focused on detecting characters~\cite{chen_character_2016}, understanding latent personas~\cite{bamman_learning_2013}, their emotions~\cite{kim_frowning_2019}, and their relationships~\cite{chaturvedi_unsupervised_2017}. Another line of work has attempted to describe characters with a set of attributes~\cite{zhang_generating_2019} or personality types~\citep{sang_mbti_2022}.
Most prior research studies characters in short stories or 
adopts relatively simplistic analysis methods (e.g.,~summaries) when it comes to long narratives.

In this paper, we focus on analyzing characters in long-form stories, a relatively understudied area that presents unique challenges not found in short stories. Firstly, long stories typically contain a large number of characters with complex relationships and interactions which have a key role in the plot. Secondly, characters in long stories are dynamic~\citep{chaturvedi_unsupervised_2017}: they develop throughout the story and their personalities, motivations, and relationships change as the plot evolves. Finally, long narratives exceed the input length that many current transformer-based architectures~\citep{vaswani_attention_2017} can process, making the problem technically challenging. 

We work towards addressing these challenges and study characters from a text-generation perspective, focusing on two tasks: (1)~\emph{character description} produces a general profile of a character (e.g.,~their actions, relationships, attributes) and (2)~\emph{character analysis} produces an in-depth interpretation of a character's personality and behaviour (e.g.,~how the character's personality develops, their motives, or the social context). The character description task has been introduced with the release of the LiSCU dataset \cite{brahman_let_2021}, which contains literary book summaries paired with human-written character descriptions. However, using summaries to describe characters significantly simplifies and restricts the task. Summaries contain limited information about the overall story, usually only the salient events, and important details are omitted. At the same time, a book summary cannot be used to describe every character of a narrative, but only those important enough to figure in the summary. 

For these reasons, our work focuses exclusively on describing characters attested in full-length books. In addition, we propose character analysis as a new task, which complements and extends character description in that it requires a more in-depth understanding of  characters. It goes beyond just describing surface-level traits, critically analyzing the character's depth, complexity, and evolution within the narrative context. Character analyses also typically explore the social, political, or historical context relevant to understanding a character and their behaviour. We show an example of these two tasks in Figure~\ref{fig:description_vs_analysis}. Additional examples can be found in Appendix~\ref{sec:data_examples}.

While previous work has made a significant effort to understand characters individually, treating them as isolated entities is a simplification.  Characters have their own arcs in a story, but they are also interconnected -- their actions, motivations, and relationships are all intertwined within the narrative \cite{Weiland:2016}.
Based on these observations, we introduce \textit{Joint Character Description}, a variation of the character description task, where the model has to generate a description for \emph{every} character sequentially. Our analysis shows that although current state-of-the-art language models can benefit from knowing all characters in a story, they struggle with joint character understanding.

Our contributions in this work are as follows:
\begin{itemize}
 \setlength\itemsep{-0.2em}
  \vspace{-0.2cm}
  \item We propose \textsc{BookWorm}, a new dataset which enables fine-grained character comprehension for long-form texts and supports the tasks of character description (in isolation and jointly) and analysis. 
  \item We establish baseline performance by training various state-of-the-art long context models, combined with different approaches to retrieving character information from long texts.
  \item  Our experiments show that retrieval-based models lead to better performance on both tasks, despite hierarchical processing \cite{chang_booookscore_2024} being the de facto approach for book summarization in the literature. 
  \item  We expose limitations in the understanding capabilities of current models whose performance degrades when they attempt to reason about characters jointly.  
\end{itemize}

\section{Related Work}

\paragraph{Narrative Structure}
Existing work has studied narratives and their plot structure, focusing primarily on summarization. Several datasets have been developed for narrative summarization. Examples include \textit{TRIPOD}~\cite{papalampidi_movie_2019}, which contains movie scripts annotated with salient scenes or turning points. \textit{NarraSum} \cite{zhao_narrasum_2022} has  summaries of movies and TV series, while \textit{BookSum} \cite{kryscinski_booksum_2021} is a collection of literary artefacts (e.g.,~novels, plays) paired with summaries.  Summarization is related to our character description and analysis tasks, but there are significant differences \citep{brahman_let_2021}. A summary captures the entire plot of a story and includes \emph{all} main characters, while a character description focuses on a \emph{single} character, their properties and actions, incorporating plot elements only when they help describe the character.

\paragraph{Character Understanding}

Some prior research has studied narratives from a character-centric perspective, focusing on a variety of tasks: the identification of character personality~\citep{sang_mbti_2022, bamman_learning_2013}, prediction of character emotions~\citep{brahman_modeling_2020} and relationships~\citep{chaturvedi_unsupervised_2017}, character detection~\citep{chen_character_2016}, grounding~\citep{liu_detecting_2023}, and the generation of character descriptions~\citep{brahman_let_2021}. Another line of research has used characters as a means to generate new stories~\citep{liu_character-centric_2020} or summaries~\citep{zhang_generating_2019} of existing narratives. However, most existing research focuses on characters of short stories and studies only specific aspects (e.g.,~relationships, emotions). In this paper, we focus on characters of full-length books (with an average length of approximately~100k tokens), and study them more holistically, from the perspective of understanding their attributes, actions, and behavior (description task) \emph{and} how are these interpreted in the context of the narrative (analysis task).

\paragraph{Character Description} Previous work~\cite{zhang_generating_2019} has found that character descriptions occur commonly in human-written story summaries, thus advocating the identification of a set of character attributes as a useful intermediate step for automatic summarization. \citet{chen_tvstorygen_2022} introduced  \emph{TVStoryGen}, a dataset aiming to generate TV episode recaps based on summaries and character descriptions. \citet{carlsson_gandalf_2021} released \emph{Gandalf}, a dataset containing descriptions paired with multiple character names, with the task of choosing the correct one. \citet{brahman_let_2021} created \emph{LiSCU}, a dataset which contains book summaries and human-written character descriptions. A part of this dataset includes full books, however, it has not been publicly released and the work focuses on describing characters from summaries, not books. 
The current paper contributes to this literature by introducing \textsc{BookWorm}, a new dataset for understanding characters based on the full-length books, without assuming that summaries are available. We also introduce the new task of character analysis, which aims to generate a more detailed account of a character's personality, motives, development, and social context. We compare our dataset with LiSCU in Section~\ref{data_analysis_subsection}.

\paragraph{Long-context Models}
Several papers have focused on alleviating the memory requirements of transformers, which increase quadratically with the input length. Sparse attention approaches like BigBird~\cite{zaheer_big_2021}, Longformer~\cite{beltagy_longformer_2020}, and Reformer~\cite{kitaev_reformer_2020} combine local windowed attention with global attention on a subset of tokens, enabling modelling of much longer sequences. Transformer-XL~\cite{dai_transformer-xl_2019} introduces segment-level recurrence as another technique for capturing longer-range dependencies.

Another line of research has sought to overcome the limitations of input length by employing retrieval-augmented generation; \citet{xu_retrieval_2023} show that retrieval can outperform long context transformers even when using shorter input. Other work~\cite{wu_recursively_2021, chang_booookscore_2024} processes long documents hierarchically by segmenting the input into shorter chunks and generating intermediate responses, which are then aggregated into a final summary. 
We propose several retrieval-augmented models for our tasks, exploring different content extraction strategies  (e.g.,~based on characters or a retrieval engine like BM25), and show they are superior to hierarchical generation.

\begin{table*}[t]
\centering
\resizebox{\textwidth}{!}{
\begin{tabular}{lcrcrrrr}
\toprule
\multirow{2}{*}{\textbf{Dataset}} & \multirow{2}{*}{\textbf{Books}} & \multirow{2}{*}{\textbf{Samples}} & \multirow{2}{*}{\textbf{Avg. Characters}} & \multicolumn{2}{c}{\textbf{Avg. Input Length}} & \multicolumn{2}{c}{\textbf{Avg. Output Length}} \\ 
 &  &  &  & \textbf{words} & \textbf{sentences} &  \textbf{words} & \textbf{sentences}\\
\midrule
BookSum & 187 & 405 &  --- &  108,477.13 & 5,195.34 & 1,151.86 & 54.84 \\
LiSCU-summary & --- & 9,499 & 5.56 & 1,022.32   & 48.82 &  184.57 & 8.56 \\
LiSCU-book & 204 & 2,052 & --- & \multicolumn{1}{c}{---}  & \multicolumn{1}{c}{---} & \multicolumn{1}{c}{---} & \multicolumn{1}{c}{---}\\
\midrule
\textsc{BookWorm} (description)  & 324 & 5,869 & 9.74 & 97,685.82 & 4,481.16 & 88.79 & 3.97 \\
\textsc{BookWorm} (analysis)  & 133 & 1,328 & 5.69 & 95,758.79  & 4,541.39 & 602.65 & \hspace*{-.2cm}25.71 \\
\bottomrule
\end{tabular}}
\caption{Satistics for \textsc{BookWorm} and comparison with related datasets (LiSCU and BookSum).  We show the total number of books, sample counts, and average length of input and output in terms of words and sentences.}
\label{data stats}
\end{table*}

\section{The \textsc{BookWorm} Dataset}

\subsection{Data Collection}

Following previous work~\cite{kryscinski_booksum_2021, brahman_let_2021}, we collect books from the Gutenberg Project\footnote{\url{https://www.gutenberg.org/}}, which contains classic books, including novels, plays, and works of poetry. To obtain character descriptions and analyses, we scrape five different websites, namely \textit{Sparknotes},  \textit{Litcharts}, \textit{Gradesaver}, \textit{Cliffnotes}, and \textit{Shmoop}.\footnote{\url{https://www.sparknotes.com/lit/}, \url{https://www.litcharts.com/}, \url{https://www.gradesaver.com/}, \url{https://www.cliffsnotes.com/}, \url{https://www.shmoop.com/}}
These websites contain complete studies of literary books, mainly for educational purposes. For our work, we use \textit{Litcharts}, \textit{Sparknotes}, \textit{Gradesaver} and \textit{Cliffsnotes} as sources for character descriptions and \textit{Sparknotes}, \textit{Shmoop} and \textit{Cliffsnotes} as sources for character analyses. For websites that are used in both tasks, there is a clear distinction between the different texts, for instance, Sparknotes contains a character list with descriptions and then more detailed and longer analyses entitled ``in-depth analysis''. For websites that are used only for description (e.g.,~Gradesaver) or analysis (e.g.,~Shmoop), we proceed to this selection after manually inspecting the corresponding text, their content, level of detail and length. To pair the books with their corresponding character descriptions and analyses, we match titles (after removing punctuation and lower-casing) and then manually verify that the authors are the same. We exclude books which belong to the genre ``philosophy'' as they do not have characters in the traditional sense. Additionally, we filter out character descriptions that are less than 30~words and character analyses that are less than 200~words to avoid short samples that do not fit the purposes of the tasks; usually, these correspond to minor characters in a story. 

We split the data (for both tasks) into train/validation/test (80/10/10) partitions based on the book titles to avoid data leakage. All books in our dataset are in the public domain and do not have any copyright restrictions. We are not allowed to redistribute data from literature websites and thus use the Web Archive\footnote{\url{http://www.web.archive.org/}} to save scraped URLs, preserving the snapshot used in our experiments. We release our corpus to encourage future work on our tasks~\footnote{\url{https://github.com/apapoudakis/BookWorm}} (see examples in Figure~\ref{fig:description_vs_analysis} and in Appendix~\ref{sec:data_examples}).

\subsection{Data Analysis}
\label{data_analysis_subsection}
We present various statistics on \textsc{BookWorm} in Table~\ref{data stats} and compare it with the related LiSCU dataset~\cite{brahman_let_2021}. For completeness, we also report statistics for BookSum\footnote{We show statistics for the book-level partition.}~\cite{kryscinski_booksum_2021}, a book summarization dataset. 

Firstly, we note that the average book length for the description and analysis tasks is approximately 95k~words, which is challenging even for current state-of-the-art transformer-based models. Additionally, we observe significant differences in task requirements; the average length of a description is 88~words, whereas the average analysis is 602~words. We have~324 and 133~unique books paired with~5,869 character descriptions and 1,328~character analyses, respectively. The LiSCU-summary partition has~9,499 samples, but contains only the summary of the story and not the full book; obtaining whole books is significantly harder than just collecting book summaries. Although the LiSCU-book partition is not publicly available,  we report numbers from the corresponding paper \cite{brahman_let_2021} in Table~\ref{data stats}. 

\begin{figure}[t]
    \centering
    \includegraphics[width=\columnwidth]{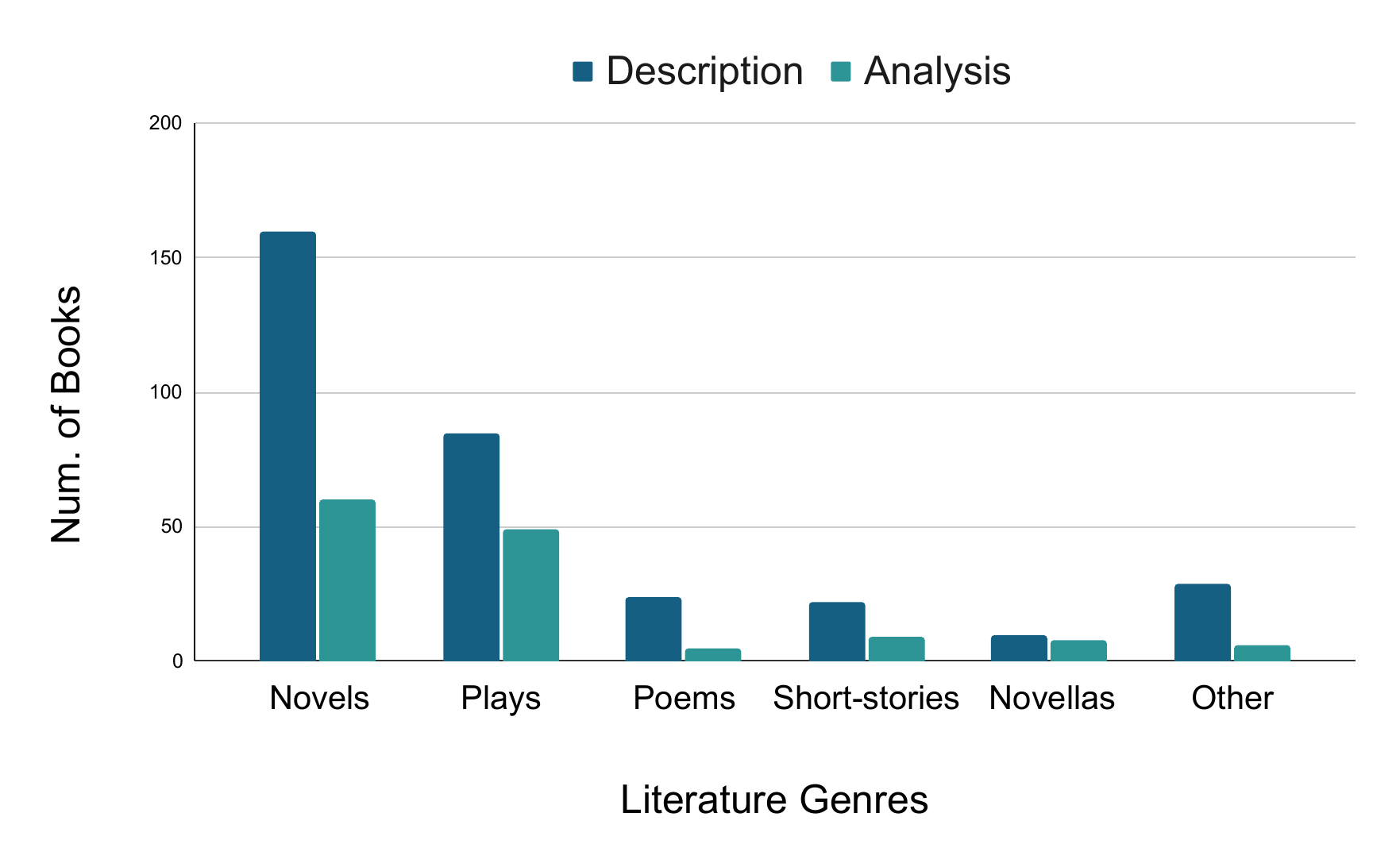}
    \caption{Distribution of genres in \textsc{BookWorm} for character description and analysis tasks.}
    \label{fig:literature_genres}
\end{figure}

Following~\citet{kryscinski_booksum_2021}, we show the literary genres represented in \textsc{BookWorm} in Figure~\ref{fig:literature_genres}. We observe similar trends across tasks: the books are mostly novels and plays, with some short stories, novellas and poetry collections; other genres (children's books, biographies and historical books) are sparsely represented.

\section{Modeling Experiments}

We conduct a series of experiments to benchmark the performance of current models and analyze their abilities across different dimensions. Initially, we explore the limits of simple extractive heuristics. Next, we evaluate an instruction-tuned model in a zero-shot setting and contrast its performance against fine-tuned models. Additionally, across our experiments, we evaluate different retrieval strategies that are generic or rely on domain-specific information and compare them with the hierarchical approach, which uses the full story.

\subsection{Extractive Heuristics}
We adopt the Lead-$k$ baseline~\cite{narayan_dont_2018}, which traditionally extracts the first~$k$
sentences from a source document. We adjust this baseline experiment to our task, extracting the first
$k$~sentences in which the character of interest is mentioned. We use the coreference model of the BookNLP library\footnote{\url{https://github.com/booknlp/booknlp}} to identify character mentions. Similarly, we define a Random baseline by randomly extracting $k$~sentences in which the character is mentioned. To define an upper bound of the extractive experiments, we develop an Extractive Oracle baseline, selecting the $k$~sentences with the highest average rouge score against the gold-standard~\cite{narayan_dont_2018}. For the description task, we set $k$~equal to four, while for the analysis task, we set~$k$ to~25, based on the average number of sentences for each task in \textsc{BookWorm}.

\subsection{Zero-shot Abstractive Models}
\label{sec:zero-shot}

All zero-shot experiments use Llama-3-8B-Instruct~\cite{dubey_Llama_2024} as a backbone model,  with an input context of~8,192 tokens.\footnote{We opted for~8k context as the model has been pre-trained with maximum sequences of this size.} As a simple baseline, we feed the model with the lead 8,192 tokens, truncating the rest of the input. 

We further develop retrieval-augmented models, following an extract-and-generate approach. In one variant, we use the statistical BM25 method~\cite{robertson1995okapi} to extract relevant context. Specifically, we use the character's name as a query and select the 80~paragraphs with the highest score. In another variant, we use a coreference model from the BookNLP library to identify character mentions and extract paragraphs in which the target character is mentioned, following prior work~\cite{brahman_let_2021, maddela_entsum_2022}. We then concatenate these paragraphs and feed them into the language model.

We compare retrieval-augmented models to a hierarchical approach in which all the book information is processed. Following previous work~\cite{wu_recursively_2021, chang_booookscore_2024}, we split the book into chunks of 8k~tokens and generate a description for each chunk. We then concatenate these intermediate descriptions and feed them to the language model, which merges them into a final description. We experimented with adding more steps to the hierarchical approach, but we did not observe an improvement, and thus only report the results of \emph{single-step} hierarchical processing in this paper. We show the prompts used for our zero-shot models in Appendix~\ref{sec:train_details}.

\subsection{Fine-tuned Abstractive Models}
\label{sec:fine-tuning}

We experiment with two architectures: the encoder-decoder LongT5-base model~\cite{guo_longt5_2022} with 16,384 tokens input context and the decoder-only Llama-3-8B-Instruct model~\cite{dubey_Llama_2024} with 8,192 tokens context length. 

Analogously to our zero-shot models, we compare the fine-tuned models to a simple baseline, which truncates the input story at the maximum length the model can process. In addition, we evaluate the two retrieval strategies mentioned earlier, namely using BM25 or the coreference model from the BookNLP library. We fully fine-tune LongT5, while for Llama-3, we do parameter efficient fine-tuning using LoRA~\cite{hu_lora_2021}. We report the hyperparameters and additional training details in Appendix~\ref{sec:train_details}.

\subsection{Generation Settings}
\label{sec:generation-settings}

We report experiments in two settings. The first setting is common in previous work \cite{brahman_let_2021} and aims to generate a description or analysis for a character in isolation. In addition, we explore an alternative formulation where we collectively describe or analyse all the characters in a story. We call this setting \emph{joint character description}. 
We also explore a variant where the model has to describe every character separately, but all the characters from the story are given as input in the prompt, so as to ensure parity of context for the two alternative task formulations. 

For the joint description setting, we employ the hierarchical approach in a zero-shot fashion as described in Section~\ref{sec:zero-shot}. We use Llama-3-70B-Instruct as the base model because we find that smaller models fail to describe the characters jointly and output descriptions for each. This is particularly problematic for books with many characters. In this case, we adjust and prompt the model to describe five characters at a time instead of all characters together. If the model still struggles to follow the required template, then we describe the characters individually. We report the prompts used for these experiments in Appendix~\ref{sec:train_details}.

\subsection{Evaluation Metrics}
Automated evaluation metrics are crucial for our task and for related book-length applications where human evaluation is extremely labor-intensive, costly, and difficult to design~\cite{krishna_longeval_2023}. As there is no single agreed-upon metric for automatically measuring character understanding, we 
evaluate output quality along different dimensions and report several complementary metrics. 

We use Rouge F1~\cite{lin_rouge_2004} against the reference descriptions as a way of assessing the informativeness of descriptions or analyses. We report Rouge-1 (unigram overlap), Rouge-2 (bigram overlap), and Rouge-L (longest common subsequence between the model output and the gold-standard description). We also report entity mention recall following prior work~\cite{bertsch_unlimiformer_2023}, which counts the percentage of named entities (e.g.,~person names, locations) present in the reference that are covered by the model output. Additionally, we use BERTScore~\citep{zhang_bertscore_2020}, which calculates token similarity using contextual embeddings instead of string matching.

As token-matching evaluation does not always correlate well with the quality of the generated text~\citep{fabbri_summeval_2021}, we also use QA-based evaluation, following existing literature~\citep{deutsch_towards_2021, fabbri_qafacteval_2022}. Specifically, we create a set of question-answer pairs based on the reference descriptions and then use the model output to answer these questions. We expect factual descriptions to correctly answer a higher percentage of questions. We first prompt GPT-3.5, asking it to generate question-answer pairs based on gold-standard descriptions. Since question-answering models are typically trained on data different from the narrative domain, such as Wikipedia passages, we fine-tune a RoBERTa-large encoder~\cite{liu_roberta_2019} using QA pairs from our dataset. We discard low-quality questions through round-trip filtering~\citep{alberti_synthetic_2019}, i.e.,~we check whether the generated questions can indeed be answered using the reference description. We employ exact match and F1~\cite{rajpurkar_squad_2016} to evaluate all QA models. We present examples of the question-answering evaluation in Appendix~\ref{sec:qa_based_eval}.

As all the above metrics are reference-based, we also use an entailment-based metric, which predicts whether the input story entails the model output. Specifically, following \citet{narayan_conditional_2022} and \citet{laban_summac_2021}, for each generated sentence, we calculate its maximum entailment score against the paragraphs of the input story. If a paragraph is longer than 512~tokens, we split it into shorter paragraphs. We also transform the entailment probability into 0 or 1 using a 0.5~threshold. Then, we calculate the average entailment score across the model output. As an entailment model for our experiments, we use T5-XXL~\cite{raffel_exploring_2020} fine-tuned on the Adversarial NLI dataset~\cite{nie_adversarial_2020}. 

Previous research has also used LLM-as-a-judge pipelines to assess the quality of generated text~\cite{mahon2024modular, min-etal-2023-factscore, song_veriscore_2024, zheng_judging_2023}. In this paper, we adopt the PRISMA metric~\cite{mahon2024modular} using a large language model to evaluate the factuality of the generated outputs. Specifically, we calculate PRISMA-precision by extracting facts from the generated output and then using the gold-standard to judge whether these facts are supported or not. Similarly, we calculate PRISMA-recall by extracting facts from the gold-standard and then using the model output to assess these facts. PRISMA-F1 is then derived from these precision and recall values. We used GPT-4o-mini to extract facts and judge their factuality. 

Additionally, we evaluate factuality across different character dimensions, by classifying the extracted facts into six distinct categories: Role (the part the character plays in the story), Relationship (connections the character has with others, e.g., friendships or family ties), Personality (the character's behavior, traits or attributes), Event (actions and decisions the character is involved in), Mental State (the character's state of mind, e.g., beliefs, intentions, and emotions), and Other Fact (any fact that does no belong to the above categories). We chose this categorization based on prior work~\cite{brahman_let_2021} and after having manually inspected examples of extracted facts. We again used GPT-4o-mini to classify facts into the above categories. To assess the reliability of the model’s classification, we conducted a human annotation process where the authors of this paper classified 200 facts extracted from character descriptions and analyses. We found a strong agreement among the annotators with a Fleiss' Kappa of 74.64. We also found that GPT-4o-mini correlates highly with the majority of the human annotators, achieving a Cohen's Kappa of 62.47. Additional details for the fact-based evaluation are in Appendix~\ref{sec:train_details}.

\subsection{Implementation Details}

All experiments were run on a single Nvidia A100 or H100 GPU, except for the joint description experiments, which used two H100s. We used pre-trained models from HuggingFace and trained our models for four epochs selecting the checkpoint with the lowest validation loss. During inference, we used sample decoding with a temperature of 0.4. We report additional implementation details in Appendix~\ref{sec:train_details}.

\section{Results}

\begin{table*}[ht]
\centering
\resizebox{\textwidth}{!}{%
\begin{tabular}{@{}p{0.5cm}lccccc|ccccc@{}}
\toprule
 & & \multicolumn{5}{c}{\textbf{Description}} & \multicolumn{5}{c}{\textbf{Analysis}} \\
& \textbf{Model} & R-1 & R-2 & R-L & EntMent  & \multicolumn{1}{c}{BS} & \multicolumn{1}{c}{R-1} &  R-2 & R-L & EntMent & BS \\
\midrule
\multirow{3}{*}{\rotatebox{90}{Heuristic}} & 

Lead & 20.64 &	2.08 & 12.23 & 24.08  & 49.33 & 25.20 &	2.22 & 10.96 & 12.89  & 45.87 \\
& Random & 19.67 & 1.65 & 11.58  & 22.62 & 49.34 & 25.00 & 2.22 & 10.91 & 11.68 & 45.82 \\
& Oracle & 34.38 & 8.97 & 21.46 & 21.46 & 50.75 & 42.16 & 14.03 & 17.86 & 21.13 & 48.74 \\

\midrule
\multirow{4}{*}{\rotatebox[origin=c]{90}{Zero-shot}}
& Llama-3 & 27.89 &	5.00 & 17.24 & 28.15 & 55.21 & 32.48 & 6.18 & 16.34 & 20.70 & 54.16\\
& \hspace{0.3cm} + BM25 & 29.69	& 5.95 & 18.28  & 31.60 & 57.29 & 33.59 & \textbf{6.68} & \textbf{16.75} & \textbf{24.22} & 54.56 \\
& \hspace{0.3cm} + coref & 29.86 &	5.98 &	18.55 & 32.27 & \textbf{57.90} & 33.36 & 6.60 & 16.60 & 22.80 &  \textbf{54.70} \\
& \hspace{0.3cm} + hier & 28.46 &	5.67 &	17.72 & 29.44  & 56.74 &  31.47	 & 6.09 & 15.69 & 23.16 & 53.79\\

\midrule
\multirow{6}{*}{\rotatebox[origin=c]{90}{Fine-tuning}} &
 LongT5 & 28.62 & 	5.53 & 17.91 & 26.73 &  54.91 &  33.03 & 6.27 & 15.38 & 11.86 & 50.01  \\
& \hspace{0.3cm}   + BM25 & 29.98 & 5.84 & 18.61 & 28.69  & 56.27 &  32.19 &  6.02 & 15.49 &  12.80 & 48.98\\
& \hspace{0.3cm}   + coref & 29.19	& 5.84 & 17.82 & 31.15  & 55.43 & 32.38 &	6.13 & 15.18 & 10.27 & 49.56 \\
& Llama-3 & 27.90 &	5.62 &	18.74 & 29.59 & 55.34 & 33.59 &	6.32 & 15.49 & 16.60 & 52.86 \\
& \hspace{0.3cm}   + BM25 & 29.78 & 6.42 & 19.67 & 34.80 & 56.93 & \textbf{34.10} &	6.53 & 15.62 & 18.09 & 52.87  \\
& \hspace{0.3cm}   + coref & \textbf{30.36} & \textbf{6.75} & \textbf{19.84} & \textbf{35.78} & 57.63 & 33.93 &	6.55 & 15.67 & 18.28 & 53.10 \\
\bottomrule
\end{tabular} 
}
\caption{Results on character description and analysis tasks on our \textsc{BookWorm} dataset. We use Rouge, entity mention recall (EntMent) and BERTScore (BS). Best model per metric is boldfaced (excluding the oracle).
}\label{tab:baselines}
\label{full_results}
\end{table*}

\begin{table}[t]
\centering
\resizebox{\columnwidth}{!}{
\begin{tabular}{lccc|ccc@{}}
\toprule
 & \multicolumn{3}{c}{\textbf{Description}} & \multicolumn{3}{c}{\textbf{Analysis}} \\
\textbf{Model} & EM & F1 & \multicolumn{1}{c}{NLI} & \multicolumn{1}{c}{EM} & F1 & NLI \\
\midrule
 LongT5 & 8.69 &	12.20 & 7.94 & 6.17	& 7.78 & 7.25\\
 \hspace{0.3cm}   + BM25 & 9.62 & 12.83 & 10.28 & 6.75 & 8.33 & 2.84\\
 \hspace{0.3cm}   + coref & 9.53	& 13.78 & 16.10 & 7.14	& 9.25 & 13.33 \\
Llama-3  & 8.16 & 12.04 & 16.87 & 5.58	& 8.39 & 14.15\\
\hspace{0.3cm}   + BM25 & 10.41 & 15.28 & 22.70 & 6.94 &	11.19 & 13.06 \\
\hspace{0.3cm}   + coref & \textbf{13.29} & \textbf{17.36} & \textbf{40.21} & \textbf{8.49}	& \textbf{11.27} & \textbf{50.85} \\
Reference  & -- & -- & 65.27 & -- & -- &  61.43 \\
\bottomrule
\end{tabular} }
\caption{Question answering and entailment-based evaluation for fine-tuned models. We report exact match (EM), F1, and natural language inference metric (NLI). Best model per metric is boldfaced.}\label{tab:qa_nli_results}
\label{qa}
\end{table}

\begin{table*}[ht]
\centering
\resizebox{\textwidth}{!}{%
\begin{tabular}{@{}lccccccc|ccccccc}
\toprule
\multirow{3}{*}{\textbf{Model}} & \multicolumn{7}{c}{\textbf{Description}} & \multicolumn{7}{c}{\textbf{Analysis}} \\
& \multicolumn{1}{c}{Role} & Relat. & Person. & Event & Mental S. & Other & \multicolumn{1}{c}{Overall} & \multicolumn{1}{c}{Role} & Relat. & Person. & Event & Mental S. & Other & Overall\\
\midrule
 LongT5  & 19.90 & 16.68 & 29.71 & 10.49 & 23.25 & 22.23 & 19.98 &  	26.49 &	17.83 &	28.63 &	13.20 &	23.52 &	19.73 & 22.54 \\ 
  \hspace{0.3cm}   + BM25 & 28.51 & 21.03 &	32.00 &	13.56 &	25.31 &	24.72 &  22.34 & 26.15 &17.70 &	24.90 &	12.19 &	18.69 &	25.06 &20.60 \\
 \hspace{0.3cm}   + coref & 33.41 &	26.43 &	32.33 &	14.53 &	25.72 &	28.14 & 25.46  & 24.29 &	18.32 &	25.15 &	11.08 &	18.88 &	20.69 & 19.72\\
 Llama-3  & 38.77 &	20.99 &	40.61 &	27.02 &	32.70 &	37.25 & 34.89 &	48.53 &	34.17 &	42.11 &	\textbf{29.57} &	35.59 &	37.60 & 36.56\\
 \hspace{0.3cm}   + BM25  & 51.88 &	38.99 &	51.84 &	34.04 &	39.80 & 	45.41 & 42.80 &	50.71 &	33.34 &	44.11 &	29.29 &	\textbf{37.35} &	37.82 & 37.20 \\
 \hspace{0.3cm}   + coref &	\textbf{53.66} &	\textbf{41.31} &	\textbf{53.50} &	\textbf{34.54} &	\textbf{43.39} &	\textbf{47.91}	& \textbf{52.68} & 	\textbf{52.12} &	\textbf{36.65} &	\textbf{46.11} &	29.48 &	36.85 &	\textbf{39.28} & \textbf{37.95} \\

\bottomrule
\end{tabular} 
 }
\caption{PRISMA evaluation results on description and analysis tasks for the fine-tuned models. We report the overall PRISMA-F1 score along with the F1 score of each subcategory (Role, Relationship, Personality, Event, Mental State and Other Fact). Best model per metric is boldfaced.}\label{tab:fact_based_results}

\end{table*}

\paragraph{There is no lead bias in book-length character understanding.} Our experimental results are summarized in Table~\ref{full_results}.  Lead-$k$ performs poorly, even though it is a strong baseline in standard summarization tasks~\citep{nallapati_abstractive_2016, narayan_dont_2018}. It achieves substantially lower scores in terms of Rouge compared to zero-shot and fine-tuned models. Random selection performs similarly, achieving marginally worse scores than the Lead baseline in both tasks. 

The extractive oracle heuristic achieves the highest Rouge scores across all experiments in both tasks. There is a bigger performance gap when it comes to the analysis task, where oracle experiment is ostensibly better, especially in Rouge-1 and Rouge-2, compared to zero-shot and fine-tuned models. This result is expected as the oracle model uses gold-standard texts to extract sentences. When considering Entity Mention recall, we observe that the Oracle model is worse at the description task than zero-shot and fine-tuned models. Interestingly, in the analysis task, while the Oracle model scores lower than zero-shot models in Entity Mention recall, it surpasses the fine-tuned models in this metric. This result demonstrates that there is still space for improvement in the way that our experiments retrieve context and use salient entities.

BERTScore results for the Oracle model are comparable to the Lead and Random heuristics, and lower compared to abstractive models. This is an expected outcome, as the Oracle fails to capture the semantic information of the reference descriptions, even if it matches the gold-standard tokens. 

\paragraph{Retrieval-augmented models perform best in both character description and analysis.}
In our zero-shot experiments, we observe that the retrieval-based methods consistently improve performance in both tasks. Specifically, the coreference approach outperforms BM25 in the description task while BM25 performs better in the analysis. The hierarchical approach improves results compared to the Lead baseline in the description task but does not match the performance of retrieval-based methods. In the analysis task, the performance is slightly worse than in the Lead experiment. We hypothesize that this occurs because retrieving relevant information is more crucial than processing the entire story for tasks like character description and analysis, which resemble query-based summarization more than generic summarization.

For our fine-tuned models, we observe trends similar to the zero-shot ones. Specifically, both the BM25 and coreference-based retrieval lead to better descriptions and analyses, with the exception of LongT5 in the analysis task, where the differences are only marginal. Llama-3 consistently outperforms LongT5 across both tasks. While fine-tuning leads to consistent improvements in the description task, this is not the case for the analysis task where fine-tuning is either comparable or inferior to the  zero-shot setting. We hypothesize that there are two reasons for this, the training samples are fewer in the analysis task and the level of data contamination is higher (see Appendix~\ref{sec:additional_experiments}). 

\paragraph{Fine-tuned Llama with coreference-based retrieval is the most faithful.}

We report QA-based and NLI-based evaluation results in Table~\ref{qa}. We focus on fine-tuned models as these performed better in most cases than zero-shot ones and extractive baselines, according to reference-based metrics (see Table~\ref{tab:baselines}). QA-based metrics reward Llama most when enhanced with coreference-based retrieval in both tasks. In general, performance improves when relevant context is retrieved in both the LongT5 and Llama models. Llama consistently outperforms LongT5, and coreference-based retrieval yields better results than extraction using~BM25.

\begin{table}[t]
    \centering
    \begin{tabular}{@{}lccc@{}}
        \toprule
        \textbf{Model} & R-L & EntMent & QA-F1 \\
        \midrule
        Separate & 17.82 & 29.92 & 14.37  \\
       \hspace{0.3cm} +  character names  & \textbf{18.36} & \textbf{31.71} & \textbf{14.63} \\
        Joint & 16.62 & 23.39 & 12.78  \\
        \bottomrule
    \end{tabular} 
    \caption{Character description results in separate and joint generation settings. We report  Rouge-L, entity mention recall (EntMent) and question answering~F1. We use the hierarchical approach with zero-shot Llama-3-70B-Instruct. Best model per metric is boldfaced.}\label{tab:separate_vs_joint}
\end{table}

The NLI metric has a clear preference for models fine-tuned on coreference-based input. In particular for Llama, we observe a large jump in entailment accuracy over BM25. Retrieving relevant context helps achieve higher entailment scores for both tasks. The only exception is the entailment accuracy of LongT5 combined with BM25 on the analysis task, where performance decreases compared to LongT5 on its own. The coreference resolution approach is consistently better than BM25.

\paragraph{Facts related to events and relationships are hard to get right.}
We report the fact-based evaluation in Table~\ref{tab:fact_based_results}. We observe that retrieval-augmented models demonstrate higher overall factuality, leading to improvements across nearly all character dimensions for both tasks. An exception is the LongT5 model for the character analysis task, where the lead baseline outperforms BM25 and coreference-based models. The coreference-based model surpasses BM25 in description and analysis, while the Llama model consistently outperforms LongT5. Across both tasks, facts related to events and character relationships are the least factual. In contrast, facts concerning a character's role and personality achieve the highest scores. Mental state and other facts perform similarly, but they fall below those related to personality and role. Our results demonstrate that models struggle with the more dynamic aspects of characters, such as events and relationships, while handling more static dimensions like role and personality more effectively. Examples of the fact-based evaluation are in Appendix~\ref{sec:qa_based_eval}. 

\paragraph{It is easier to talk about one character than about many.} Table~\ref{tab:separate_vs_joint} presents results in the two generation settings: joint and separate character description.\footnote{We do not perform joint experiments for the analysis task, as this would be extremely challenging.} For this comparison, we employ the hierarchical method in a zero-shot setting with Llama-3-70B-instruct (see Section~\ref{sec:generation-settings}), as we observed that smaller models could not follow instructions for the joint task. 

The model generally struggles with the joint task, performing consistently worse across metrics compared to describing each character individually. As we can see in Table~\ref{tab:separate_vs_joint}, the model benefits from having a list of the characters in the story. We observe performance gains across metrics when character names are included in the input. We believe the joint description task is too difficult for the model which is now required to understand the story from beginning to end instead of being able to focus on a single character. Aside from understanding being harder, generation is also more challenging, as the output is quite long in this setting. Even Llama-3-70B struggles to describe \emph{all} the characters. Examples of generated outputs are in Appendix~\ref{sec:qa_based_eval}.

\section{Discussion}

Our experiments underline the importance of retrieving relevant context; we found that even simple methods such as statistical retrieval with BM25 or coreference-based retrieval lead to consistent improvements in all our experiments. Notably, while the hierarchical approach is considered state-of-the-art for book summarization, our experiments revealed it performs worse than retrieval in both description and the analysis tasks. The difference between retrieval-based and hierarchical approaches is significant even for character analysis. One might conjecture that processing the whole book would be beneficial, however, this is not corroborated by our results.   

 Until now, characters have been studied separately in the literature, which is a significant simplification. Our experiments with joint understanding of characters show that a separate description model can benefit from knowing all the different characters in a story if we list them in the initial prompt. However, our experiments also demonstrate that models struggle to understand characters jointly, having to ``read'' a book multiple times to be able to describe each character separately.

\section{Conclusions}

In this work, we created \textsc{BookWorm}, a new dataset which contains books from the Gutenberg project and human-written character descriptions and character analyses from literature websites. Character descriptions are short and factual, while character analyses are longer; they explore the motives, personality, and development of a character and often also comment on the social, historical, or political context.
We established a set of baselines using simple extractive heuristics as well as retrieval-based and hierarchical long-context models, in both zero-shot and fine-tuning settings. Our experiments highlight the importance of retrieving relevant context, which leads to consistent improvements and outperforms hierarchical methods. 

We hope our findings will inspire future research on character analysis, and text generation from long documents more generally.
We plan to develop a better suited model for the joint character description task, by keeping track of characters and their relations as the narrative evolves.  Evaluation is another avenue for future work. Entailment-based metrics are good indicators of model performance for retrieval-augmented approaches, but are computationally challenging for book-length inputs. Question-answering evaluation helps assess the factuality of the generated text but is constrained by the availability of references, and can be too punitive (in cases where model predictions have no lexical overlap with the reference). We used a LLM as a judge to perform a fact-based evaluation and gain a deeper understanding of the factuality of the different character dimensions. However, these results are again solely based on reference texts. In the future, there is a need to explore evaluation metrics that consider the full input text and are at the same time efficient and scalable.

\section*{Limitations}

Our dataset contains publicly available books discussed widely across multiple sources (reviews, critical essays, literary commentaries, etc). Even if models have not been trained on the description and analysis tasks, it is likely that they have been exposed to these literary texts or related information during pre-training. To mitigate the risk of data contamination, future work should consider using books that are not publicly available. 

In this paper, we relied on automatic metrics such as Rouge, QA-based evaluation, entailment and fact-based scores, entity mention recall and BERTScore to assess the quality of generated descriptions and analyses; however, the majority of these metrics are reference-based and do not consider book-length input to evaluate different aspects of model output. Future work could focus on reference-free evaluation metrics and efficient methods to conduct human-based evaluation. 

Moreover, current language models used in this paper do not provide any explanation about the generated text. Future research could focus more on attributable language models that generate text pointing to specific parts of the input. This would also mitigate the difficulty of conducting human evaluation, especially for tasks like character description or analysis, where many responses can be produced, but it is important to evaluate whether they are faithful. 

Our work considers simple retrieval-based strategies such as BM25 and the use of an off-the-self coreference model. A natural next step would be to explicitly train a retriever model for the two tasks in the \textsc{BookWorm} dataset. Finally, we present experiments with only one type of hierarchical model in the joint character description setting. Follow-on work could study this setting in more depth.

\section*{Acknowledgements}
This work was supported in part by the UKRI Centre for Doctoral Training in Natural Language Processing, funded by the UKRI (grant EP/S022481/1) and the University of Edinburgh, School of Informatics and School of Philosophy, Psychology \& Language Sciences. Lapata gratefully acknowledges the support of the UK Engineering and Physical Sciences Research Council (grant EP/L016427/1).

\bibliography{references, custom}

\begin{thebibliography}{52}
\providecommand{\natexlab}[1]{#1}

\bibitem[{Alberti et~al.(2019)Alberti, Andor, Pitler, Devlin, and Collins}]{alberti_synthetic_2019}
Chris Alberti, Daniel Andor, Emily Pitler, Jacob Devlin, and Michael Collins. 2019.
\newblock \href {http://arxiv.org/abs/1906.05416} {Synthetic {QA} {Corpora} {Generation} with {Roundtrip} {Consistency}}.
\newblock \emph{arXiv preprint}.
\newblock ArXiv:1906.05416 [cs].

\bibitem[{Bamman et~al.(2013)Bamman, O'Connor, and Smith}]{bamman_learning_2013}
David Bamman, Brendan O'Connor, and Noah~A. Smith. 2013.
\newblock \href {https://aclanthology.org/P13-1035} {Learning {Latent} {Personas} of {Film} {Characters}}.
\newblock In \emph{Proceedings of the 51st {Annual} {Meeting} of the {Association} for {Computational} {Linguistics} ({Volume} 1: {Long} {Papers})}, pages 352--361, Sofia, Bulgaria. Association for Computational Linguistics.

\bibitem[{Beltagy et~al.(2020)Beltagy, Peters, and Cohan}]{beltagy_longformer_2020}
Iz~Beltagy, Matthew~E. Peters, and Arman Cohan. 2020.
\newblock \href {http://arxiv.org/abs/2004.05150} {Longformer: {The} {Long}-{Document} {Transformer}}.
\newblock \emph{arXiv preprint}.
\newblock ArXiv:2004.05150 [cs].

\bibitem[{Bertsch et~al.(2023)Bertsch, Alon, Neubig, and Gormley}]{bertsch_unlimiformer_2023}
Amanda Bertsch, Uri Alon, Graham Neubig, and Matthew~R. Gormley. 2023.
\newblock \href {http://arxiv.org/abs/2305.01625} {Unlimiformer: {Long}-{Range} {Transformers} with {Unlimited} {Length} {Input}}.
\newblock \emph{arXiv preprint}.
\newblock ArXiv:2305.01625 [cs].

\bibitem[{Brahman and Chaturvedi(2020)}]{brahman_modeling_2020}
Faeze Brahman and Snigdha Chaturvedi. 2020.
\newblock \href {https://doi.org/10.18653/v1/2020.emnlp-main.426} {Modeling {Protagonist} {Emotions} for {Emotion}-{Aware} {Storytelling}}.
\newblock In \emph{Proceedings of the 2020 {Conference} on {Empirical} {Methods} in {Natural} {Language} {Processing} ({EMNLP})}, pages 5277--5294, Online. Association for Computational Linguistics.

\bibitem[{Brahman et~al.(2021)Brahman, Huang, Tafjord, Zhao, Sachan, and Chaturvedi}]{brahman_let_2021}
Faeze Brahman, Meng Huang, Oyvind Tafjord, Chao Zhao, Mrinmaya Sachan, and Snigdha Chaturvedi. 2021.
\newblock \href {http://arxiv.org/abs/2109.05438} {"{Let} {Your} {Characters} {Tell} {Their} {Story}": {A} {Dataset} for {Character}-{Centric} {Narrative} {Understanding}}.
\newblock \emph{arXiv preprint}.
\newblock ArXiv:2109.05438 [cs].

\bibitem[{Carlsson et~al.(2021)Carlsson, Sahlgren, Olsson, and Cuba~Gyllensten}]{carlsson_gandalf_2021}
Fredrik Carlsson, Magnus Sahlgren, Fredrik Olsson, and Amaru Cuba~Gyllensten. 2021.
\newblock \href {https://doi.org/10.18653/v1/2021.mrqa-1.13} {{GANDALF}: a {General} {Character} {Name} {Description} {Dataset} for {Long} {Fiction}}.
\newblock In \emph{Proceedings of the 3rd {Workshop} on {Machine} {Reading} for {Question} {Answering}}, pages 119--132, Punta Cana, Dominican Republic. Association for Computational Linguistics.

\bibitem[{Chang et~al.(2024)Chang, Lo, Goyal, and Iyyer}]{chang_booookscore_2024}
Yapei Chang, Kyle Lo, Tanya Goyal, and Mohit Iyyer. 2024.
\newblock \href {http://arxiv.org/abs/2310.00785} {{BooookScore}: {A} systematic exploration of book-length summarization in the era of {LLMs}}.
\newblock \emph{arXiv preprint}.
\newblock ArXiv:2310.00785 [cs].

\bibitem[{Chaturvedi et~al.(2017)Chaturvedi, Iyyer, and Iii}]{chaturvedi_unsupervised_2017}
Snigdha Chaturvedi, Mohit Iyyer, and Hal~Daume Iii. 2017.
\newblock \href {https://doi.org/10.1609/aaai.v31i1.10982} {Unsupervised {Learning} of {Evolving} {Relationships} {Between} {Literary} {Characters}}.
\newblock \emph{Proceedings of the AAAI Conference on Artificial Intelligence}, 31(1).
\newblock Number: 1.

\bibitem[{Chen and Gimpel(2022)}]{chen_tvstorygen_2022}
Mingda Chen and Kevin Gimpel. 2022.
\newblock \href {http://arxiv.org/abs/2109.08833} {{TVStoryGen}: {A} {Dataset} for {Generating} {Stories} with {Character} {Descriptions}}.
\newblock \emph{arXiv preprint}.
\newblock ArXiv:2109.08833 [cs].

\bibitem[{Chen and Choi(2016)}]{chen_character_2016}
Yu-Hsin Chen and Jinho~D. Choi. 2016.
\newblock \href {https://doi.org/10.18653/v1/W16-3612} {Character {Identification} on {Multiparty} {Conversation}: {Identifying} {Mentions} of {Characters} in {TV} {Shows}}.
\newblock In \emph{Proceedings of the 17th {Annual} {Meeting} of the {Special} {Interest} {Group} on {Discourse} and {Dialogue}}, pages 90--100, Los Angeles. Association for Computational Linguistics.

\bibitem[{Dai et~al.(2019)Dai, Yang, Yang, Carbonell, Le, and Salakhutdinov}]{dai_transformer-xl_2019}
Zihang Dai, Zhilin Yang, Yiming Yang, Jaime Carbonell, Quoc~V. Le, and Ruslan Salakhutdinov. 2019.
\newblock \href {http://arxiv.org/abs/1901.02860} {Transformer-{XL}: {Attentive} {Language} {Models} {Beyond} a {Fixed}-{Length} {Context}}.
\newblock \emph{arXiv preprint}.
\newblock ArXiv:1901.02860 [cs, stat].

\bibitem[{Deutsch et~al.(2021)Deutsch, Bedrax-Weiss, and Roth}]{deutsch_towards_2021}
Daniel Deutsch, Tania Bedrax-Weiss, and Dan Roth. 2021.
\newblock \href {http://arxiv.org/abs/2010.00490} {Towards {Question}-{Answering} as an {Automatic} {Metric} for {Evaluating} the {Content} {Quality} of a {Summary}}.
\newblock \emph{arXiv preprint}.
\newblock ArXiv:2010.00490 [cs].

\bibitem[{Dubey et~al.(2024)Dubey, Jauhri, Pandey, Kadian, Al-Dahle, Letman, Mathur, Schelten, Yang, Fan, Goyal, Hartshorn, Yang, Mitra, Sravankumar, Korenev, Hinsvark, Rao, Zhang, Rodriguez, Gregerson, Spataru, Roziere, Biron, Tang, Chern, Caucheteux, Nayak, Bi, Marra, McConnell, Keller, Touret, Wu, Wong, Ferrer, Nikolaidis, Allonsius, Song, Pintz, Livshits, Esiobu, Choudhary, Mahajan, Garcia-Olano, Perino, Hupkes, Lakomkin, AlBadawy, Lobanova, Dinan, Smith, Radenovic, Zhang, Synnaeve, Lee, Anderson, Nail, Mialon, Pang, Cucurell, Nguyen, Korevaar, Xu, Touvron, Zarov, Ibarra, Kloumann, Misra, Evtimov, Copet, Lee, Geffert, Vranes, Park, Mahadeokar, Shah, van~der Linde, Billock, Hong, Lee, Fu, Chi, Huang, Liu, Wang, Yu, Bitton, Spisak, Park, Rocca, Johnstun, Saxe, Jia, Alwala, Upasani, Plawiak, Li, Heafield, Stone, El-Arini, Iyer, Malik, Chiu, Bhalla, Rantala-Yeary, van~der Maaten, Chen, Tan, Jenkins, Martin, Madaan, Malo, Blecher, Landzaat, de~Oliveira, Muzzi, Pasupuleti, Singh, Paluri, Kardas, Oldham, Rita,
  Pavlova, Kambadur, Lewis, Si, Singh, Hassan, Goyal, Torabi, Bashlykov, Bogoychev, Chatterji, Duchenne, Çelebi, Alrassy, Zhang, Li, Vasic, Weng, Bhargava, Dubal, Krishnan, Koura, Xu, He, Dong, Srinivasan, Ganapathy, Calderer, Cabral, Stojnic, Raileanu, Girdhar, Patel, Sauvestre, Polidoro, Sumbaly, Taylor, Silva, Hou, Wang, Hosseini, Chennabasappa, Singh, Bell, Kim, Edunov, Nie, Narang, Raparthy, Shen, Wan, Bhosale, Zhang, Vandenhende, Batra, Whitman, Sootla, Collot, Gururangan, Borodinsky, Herman, Fowler, Sheasha, Georgiou, Scialom, Speckbacher, Mihaylov, Xiao, Karn, Goswami, Gupta, Ramanathan, Kerkez, Gonguet, Do, Vogeti, Petrovic, Chu, Xiong, Fu, Meers, Martinet, Wang, Tan, Xie, Jia, Wang, Goldschlag, Gaur, Babaei, Wen, Song, Zhang, Li, Mao, Coudert, Yan, Chen, Papakipos, Singh, Grattafiori, Jain, Kelsey, Shajnfeld, Gangidi, Victoria, Goldstand, Menon, Sharma, Boesenberg, Vaughan, Baevski, Feinstein, Kallet, Sangani, Yunus, Lupu, Alvarado, Caples, Gu, Ho, Poulton, Ryan, Ramchandani, Franco, Saraf,
  Chowdhury, Gabriel, Bharambe, Eisenman, Yazdan, James, Maurer, Leonhardi, Huang, Loyd, De~Paola, Paranjape, Liu, Wu, Ni, Hancock, Wasti, Spence, Stojkovic, Gamido, Montalvo, Parker, Burton, Mejia, Wang, Kim, Zhou, Hu, Chu, Cai, Tindal, Feichtenhofer, Civin, Beaty, Kreymer, Li, Wyatt, Adkins, Xu, Testuggine, David, Parikh, Liskovich, Foss, Wang, Le, Holland, Dowling, Jamil, Montgomery, Presani, Hahn, Wood, Brinkman, Arcaute, Dunbar, Smothers, Sun, Kreuk, Tian, Ozgenel, Caggioni, Guzmán, Kanayet, Seide, Florez, Schwarz, Badeer, Swee, Halpern, Thattai, Herman, Sizov, Guangyi, Zhang, Lakshminarayanan, Shojanazeri, Zou, Wang, Zha, Habeeb, Rudolph, Suk, Aspegren, Goldman, Damlaj, Molybog, Tufanov, Veliche, Gat, Weissman, Geboski, Kohli, Asher, Gaya, Marcus, Tang, Chan, Zhen, Reizenstein, Teboul, Zhong, Jin, Yang, Cummings, Carvill, Shepard, McPhie, Torres, Ginsburg, Wang, Wu, U, Saxena, Prasad, Khandelwal, Zand, Matosich, Veeraraghavan, Michelena, Li, Huang, Chawla, Lakhotia, Huang, Chen, Garg, A, Silva, Bell,
  Zhang, Guo, Yu, Moshkovich, Wehrstedt, Khabsa, Avalani, Bhatt, Tsimpoukelli, Mankus, Hasson, Lennie, Reso, Groshev, Naumov, Lathi, Keneally, Seltzer, Valko, Restrepo, Patel, Vyatskov, Samvelyan, Clark, Macey, Wang, Hermoso, Metanat, Rastegari, Bansal, Santhanam, Parks, White, Bawa, Singhal, Egebo, Usunier, Laptev, Dong, Zhang, Cheng, Chernoguz, Hart, Salpekar, Kalinli, Kent, Parekh, Saab, Balaji, Rittner, Bontrager, Roux, Dollar, Zvyagina, Ratanchandani, Yuvraj, Liang, Alao, Rodriguez, Ayub, Murthy, Nayani, Mitra, Li, Hogan, Battey, Wang, Maheswari, Howes, Rinott, Bondu, Datta, Chugh, Hunt, Dhillon, Sidorov, Pan, Verma, Yamamoto, Ramaswamy, Lindsay, Lindsay, Feng, Lin, Zha, Shankar, Zhang, Zhang, Wang, Agarwal, Sajuyigbe, Chintala, Max, Chen, Kehoe, Satterfield, Govindaprasad, Gupta, Cho, Virk, Subramanian, Choudhury, Goldman, Remez, Glaser, Best, Kohler, Robinson, Li, Zhang, Matthews, Chou, Shaked, Vontimitta, Ajayi, Montanez, Mohan, Kumar, Mangla, Albiero, Ionescu, Poenaru, Mihailescu, Ivanov, Li, Wang,
  Jiang, Bouaziz, Constable, Tang, Wang, Wu, Wang, Xia, Wu, Gao, Chen, Hu, Jia, Qi, Li, Zhang, Zhang, Adi, Nam, Yu, Wang, Hao, Qian, He, Rait, DeVito, Rosnbrick, Wen, Yang, and Zhao}]{dubey_Llama_2024}
Abhimanyu Dubey, Abhinav Jauhri, Abhinav Pandey, Abhishek Kadian, Ahmad Al-Dahle, Aiesha Letman, Akhil Mathur, Alan Schelten, Amy Yang, Angela Fan, Anirudh Goyal, Anthony Hartshorn, Aobo Yang, Archi Mitra, Archie Sravankumar, Artem Korenev, Arthur Hinsvark, Arun Rao, Aston Zhang, Aurelien Rodriguez, Austen Gregerson, Ava Spataru, Baptiste Roziere, Bethany Biron, Binh Tang, Bobbie Chern, Charlotte Caucheteux, Chaya Nayak, Chloe Bi, Chris Marra, Chris McConnell, Christian Keller, Christophe Touret, Chunyang Wu, Corinne Wong, Cristian~Canton Ferrer, Cyrus Nikolaidis, Damien Allonsius, Daniel Song, Danielle Pintz, Danny Livshits, David Esiobu, Dhruv Choudhary, Dhruv Mahajan, Diego Garcia-Olano, Diego Perino, Dieuwke Hupkes, Egor Lakomkin, Ehab AlBadawy, Elina Lobanova, Emily Dinan, Eric~Michael Smith, Filip Radenovic, Frank Zhang, Gabriel Synnaeve, Gabrielle Lee, Georgia~Lewis Anderson, Graeme Nail, Gregoire Mialon, Guan Pang, Guillem Cucurell, Hailey Nguyen, Hannah Korevaar, Hu~Xu, Hugo Touvron, Iliyan Zarov,
  Imanol~Arrieta Ibarra, Isabel Kloumann, Ishan Misra, Ivan Evtimov, Jade Copet, Jaewon Lee, Jan Geffert, Jana Vranes, Jason Park, Jay Mahadeokar, Jeet Shah, Jelmer van~der Linde, Jennifer Billock, Jenny Hong, Jenya Lee, Jeremy Fu, Jianfeng Chi, Jianyu Huang, Jiawen Liu, Jie Wang, Jiecao Yu, Joanna Bitton, Joe Spisak, Jongsoo Park, Joseph Rocca, Joshua Johnstun, Joshua Saxe, Junteng Jia, Kalyan~Vasuden Alwala, Kartikeya Upasani, Kate Plawiak, Ke~Li, Kenneth Heafield, Kevin Stone, Khalid El-Arini, Krithika Iyer, Kshitiz Malik, Kuenley Chiu, Kunal Bhalla, Lauren Rantala-Yeary, Laurens van~der Maaten, Lawrence Chen, Liang Tan, Liz Jenkins, Louis Martin, Lovish Madaan, Lubo Malo, Lukas Blecher, Lukas Landzaat, Luke de~Oliveira, Madeline Muzzi, Mahesh Pasupuleti, Mannat Singh, Manohar Paluri, Marcin Kardas, Mathew Oldham, Mathieu Rita, Maya Pavlova, Melanie Kambadur, Mike Lewis, Min Si, Mitesh~Kumar Singh, Mona Hassan, Naman Goyal, Narjes Torabi, Nikolay Bashlykov, Nikolay Bogoychev, Niladri Chatterji, Olivier
  Duchenne, Onur Çelebi, Patrick Alrassy, Pengchuan Zhang, Pengwei Li, Petar Vasic, Peter Weng, Prajjwal Bhargava, Pratik Dubal, Praveen Krishnan, Punit~Singh Koura, Puxin Xu, Qing He, Qingxiao Dong, Ragavan Srinivasan, Raj Ganapathy, Ramon Calderer, Ricardo~Silveira Cabral, Robert Stojnic, Roberta Raileanu, Rohit Girdhar, Rohit Patel, Romain Sauvestre, Ronnie Polidoro, Roshan Sumbaly, Ross Taylor, Ruan Silva, Rui Hou, Rui Wang, Saghar Hosseini, Sahana Chennabasappa, Sanjay Singh, Sean Bell, Seohyun~Sonia Kim, Sergey Edunov, Shaoliang Nie, Sharan Narang, Sharath Raparthy, Sheng Shen, Shengye Wan, Shruti Bhosale, Shun Zhang, Simon Vandenhende, Soumya Batra, Spencer Whitman, Sten Sootla, Stephane Collot, Suchin Gururangan, Sydney Borodinsky, Tamar Herman, Tara Fowler, Tarek Sheasha, Thomas Georgiou, Thomas Scialom, Tobias Speckbacher, Todor Mihaylov, Tong Xiao, Ujjwal Karn, Vedanuj Goswami, Vibhor Gupta, Vignesh Ramanathan, Viktor Kerkez, Vincent Gonguet, Virginie Do, Vish Vogeti, Vladan Petrovic, Weiwei Chu,
  Wenhan Xiong, Wenyin Fu, Whitney Meers, Xavier Martinet, Xiaodong Wang, Xiaoqing~Ellen Tan, Xinfeng Xie, Xuchao Jia, Xuewei Wang, Yaelle Goldschlag, Yashesh Gaur, Yasmine Babaei, Yi~Wen, Yiwen Song, Yuchen Zhang, Yue Li, Yuning Mao, Zacharie~Delpierre Coudert, Zheng Yan, Zhengxing Chen, Zoe Papakipos, Aaditya Singh, Aaron Grattafiori, Abha Jain, Adam Kelsey, Adam Shajnfeld, Adithya Gangidi, Adolfo Victoria, Ahuva Goldstand, Ajay Menon, Ajay Sharma, Alex Boesenberg, Alex Vaughan, Alexei Baevski, Allie Feinstein, Amanda Kallet, Amit Sangani, Anam Yunus, Andrei Lupu, Andres Alvarado, Andrew Caples, Andrew Gu, Andrew Ho, Andrew Poulton, Andrew Ryan, Ankit Ramchandani, Annie Franco, Aparajita Saraf, Arkabandhu Chowdhury, Ashley Gabriel, Ashwin Bharambe, Assaf Eisenman, Azadeh Yazdan, Beau James, Ben Maurer, Benjamin Leonhardi, Bernie Huang, Beth Loyd, Beto De~Paola, Bhargavi Paranjape, Bing Liu, Bo~Wu, Boyu Ni, Braden Hancock, Bram Wasti, Brandon Spence, Brani Stojkovic, Brian Gamido, Britt Montalvo, Carl
  Parker, Carly Burton, Catalina Mejia, Changhan Wang, Changkyu Kim, Chao Zhou, Chester Hu, Ching-Hsiang Chu, Chris Cai, Chris Tindal, Christoph Feichtenhofer, Damon Civin, Dana Beaty, Daniel Kreymer, Daniel Li, Danny Wyatt, David Adkins, David Xu, Davide Testuggine, Delia David, Devi Parikh, Diana Liskovich, Didem Foss, Dingkang Wang, Duc Le, Dustin Holland, Edward Dowling, Eissa Jamil, Elaine Montgomery, Eleonora Presani, Emily Hahn, Emily Wood, Erik Brinkman, Esteban Arcaute, Evan Dunbar, Evan Smothers, Fei Sun, Felix Kreuk, Feng Tian, Firat Ozgenel, Francesco Caggioni, Francisco Guzmán, Frank Kanayet, Frank Seide, Gabriela~Medina Florez, Gabriella Schwarz, Gada Badeer, Georgia Swee, Gil Halpern, Govind Thattai, Grant Herman, Grigory Sizov, Guangyi, Zhang, Guna Lakshminarayanan, Hamid Shojanazeri, Han Zou, Hannah Wang, Hanwen Zha, Haroun Habeeb, Harrison Rudolph, Helen Suk, Henry Aspegren, Hunter Goldman, Ibrahim Damlaj, Igor Molybog, Igor Tufanov, Irina-Elena Veliche, Itai Gat, Jake Weissman, James
  Geboski, James Kohli, Japhet Asher, Jean-Baptiste Gaya, Jeff Marcus, Jeff Tang, Jennifer Chan, Jenny Zhen, Jeremy Reizenstein, Jeremy Teboul, Jessica Zhong, Jian Jin, Jingyi Yang, Joe Cummings, Jon Carvill, Jon Shepard, Jonathan McPhie, Jonathan Torres, Josh Ginsburg, Junjie Wang, Kai Wu, Kam~Hou U, Karan Saxena, Karthik Prasad, Kartikay Khandelwal, Katayoun Zand, Kathy Matosich, Kaushik Veeraraghavan, Kelly Michelena, Keqian Li, Kun Huang, Kunal Chawla, Kushal Lakhotia, Kyle Huang, Lailin Chen, Lakshya Garg, Lavender A, Leandro Silva, Lee Bell, Lei Zhang, Liangpeng Guo, Licheng Yu, Liron Moshkovich, Luca Wehrstedt, Madian Khabsa, Manav Avalani, Manish Bhatt, Maria Tsimpoukelli, Martynas Mankus, Matan Hasson, Matthew Lennie, Matthias Reso, Maxim Groshev, Maxim Naumov, Maya Lathi, Meghan Keneally, Michael~L. Seltzer, Michal Valko, Michelle Restrepo, Mihir Patel, Mik Vyatskov, Mikayel Samvelyan, Mike Clark, Mike Macey, Mike Wang, Miquel~Jubert Hermoso, Mo~Metanat, Mohammad Rastegari, Munish Bansal, Nandhini
  Santhanam, Natascha Parks, Natasha White, Navyata Bawa, Nayan Singhal, Nick Egebo, Nicolas Usunier, Nikolay~Pavlovich Laptev, Ning Dong, Ning Zhang, Norman Cheng, Oleg Chernoguz, Olivia Hart, Omkar Salpekar, Ozlem Kalinli, Parkin Kent, Parth Parekh, Paul Saab, Pavan Balaji, Pedro Rittner, Philip Bontrager, Pierre Roux, Piotr Dollar, Polina Zvyagina, Prashant Ratanchandani, Pritish Yuvraj, Qian Liang, Rachad Alao, Rachel Rodriguez, Rafi Ayub, Raghotham Murthy, Raghu Nayani, Rahul Mitra, Raymond Li, Rebekkah Hogan, Robin Battey, Rocky Wang, Rohan Maheswari, Russ Howes, Ruty Rinott, Sai~Jayesh Bondu, Samyak Datta, Sara Chugh, Sara Hunt, Sargun Dhillon, Sasha Sidorov, Satadru Pan, Saurabh Verma, Seiji Yamamoto, Sharadh Ramaswamy, Shaun Lindsay, Shaun Lindsay, Sheng Feng, Shenghao Lin, Shengxin~Cindy Zha, Shiva Shankar, Shuqiang Zhang, Shuqiang Zhang, Sinong Wang, Sneha Agarwal, Soji Sajuyigbe, Soumith Chintala, Stephanie Max, Stephen Chen, Steve Kehoe, Steve Satterfield, Sudarshan Govindaprasad, Sumit Gupta,
  Sungmin Cho, Sunny Virk, Suraj Subramanian, Sy~Choudhury, Sydney Goldman, Tal Remez, Tamar Glaser, Tamara Best, Thilo Kohler, Thomas Robinson, Tianhe Li, Tianjun Zhang, Tim Matthews, Timothy Chou, Tzook Shaked, Varun Vontimitta, Victoria Ajayi, Victoria Montanez, Vijai Mohan, Vinay~Satish Kumar, Vishal Mangla, Vítor Albiero, Vlad Ionescu, Vlad Poenaru, Vlad~Tiberiu Mihailescu, Vladimir Ivanov, Wei Li, Wenchen Wang, Wenwen Jiang, Wes Bouaziz, Will Constable, Xiaocheng Tang, Xiaofang Wang, Xiaojian Wu, Xiaolan Wang, Xide Xia, Xilun Wu, Xinbo Gao, Yanjun Chen, Ye~Hu, Ye~Jia, Ye~Qi, Yenda Li, Yilin Zhang, Ying Zhang, Yossi Adi, Youngjin Nam, Yu, Wang, Yuchen Hao, Yundi Qian, Yuzi He, Zach Rait, Zachary DeVito, Zef Rosnbrick, Zhaoduo Wen, Zhenyu Yang, and Zhiwei Zhao. 2024.
\newblock \href {http://arxiv.org/abs/2407.21783} {The {Llama} 3 {Herd} of {Models}}.
\newblock \emph{arXiv preprint}.
\newblock ArXiv:2407.21783 [cs].

\bibitem[{Fabbri et~al.(2021)Fabbri, Kryściński, McCann, Xiong, Socher, and Radev}]{fabbri_summeval_2021}
Alexander~R. Fabbri, Wojciech Kryściński, Bryan McCann, Caiming Xiong, Richard Socher, and Dragomir Radev. 2021.
\newblock \href {http://arxiv.org/abs/2007.12626} {{SummEval}: {Re}-evaluating {Summarization} {Evaluation}}.
\newblock \emph{arXiv preprint}.
\newblock ArXiv:2007.12626 [cs].

\bibitem[{Fabbri et~al.(2022)Fabbri, Wu, Liu, and Xiong}]{fabbri_qafacteval_2022}
Alexander~R. Fabbri, Chien-Sheng Wu, Wenhao Liu, and Caiming Xiong. 2022.
\newblock \href {http://arxiv.org/abs/2112.08542} {{QAFactEval}: {Improved} {QA}-{Based} {Factual} {Consistency} {Evaluation} for {Summarization}}.
\newblock \emph{arXiv preprint}.
\newblock ArXiv:2112.08542 [cs].

\bibitem[{Guo et~al.(2022)Guo, Ainslie, Uthus, Ontanon, Ni, Sung, and Yang}]{guo_longt5_2022}
Mandy Guo, Joshua Ainslie, David Uthus, Santiago Ontanon, Jianmo Ni, Yun-Hsuan Sung, and Yinfei Yang. 2022.
\newblock \href {http://arxiv.org/abs/2112.07916} {{LongT5}: {Efficient} {Text}-{To}-{Text} {Transformer} for {Long} {Sequences}}.
\newblock \emph{arXiv preprint}.
\newblock ArXiv:2112.07916 [cs].

\bibitem[{He et~al.(2021)He, Liu, Gao, and Chen}]{he_deberta_2021}
Pengcheng He, Xiaodong Liu, Jianfeng Gao, and Weizhu Chen. 2021.
\newblock \href {http://arxiv.org/abs/2006.03654} {{DeBERTa}: {Decoding}-enhanced {BERT} with {Disentangled} {Attention}}.
\newblock \emph{arXiv preprint}.
\newblock ArXiv:2006.03654 [cs].

\bibitem[{Hu et~al.(2021)Hu, Shen, Wallis, Allen-Zhu, Li, Wang, Wang, and Chen}]{hu_lora_2021}
Edward~J. Hu, Yelong Shen, Phillip Wallis, Zeyuan Allen-Zhu, Yuanzhi Li, Shean Wang, Lu~Wang, and Weizhu Chen. 2021.
\newblock \href {http://arxiv.org/abs/2106.09685} {{LoRA}: {Low}-{Rank} {Adaptation} of {Large} {Language} {Models}}.
\newblock \emph{arXiv preprint}.
\newblock ArXiv:2106.09685 [cs].

\bibitem[{Kim and Klinger(2019)}]{kim_frowning_2019}
Evgeny Kim and Roman Klinger. 2019.
\newblock \href {https://doi.org/10.18653/v1/N19-1067} {Frowning {Frodo}, {Wincing} {Leia}, and a {Seriously} {Great} {Friendship}: {Learning} to {Classify} {Emotional} {Relationships} of {Fictional} {Characters}}.
\newblock In \emph{Proceedings of the 2019 {Conference} of the {North} {American} {Chapter} of the {Association} for {Computational} {Linguistics}: {Human} {Language} {Technologies}, {Volume} 1 ({Long} and {Short} {Papers})}, pages 647--653, Minneapolis, Minnesota. Association for Computational Linguistics.

\bibitem[{Kitaev et~al.(2020)Kitaev, Kaiser, and Levskaya}]{kitaev_reformer_2020}
Nikita Kitaev, Łukasz Kaiser, and Anselm Levskaya. 2020.
\newblock \href {http://arxiv.org/abs/2001.04451} {Reformer: {The} {Efficient} {Transformer}}.
\newblock \emph{arXiv preprint}.
\newblock ArXiv:2001.04451.

\bibitem[{Krishna et~al.(2023)Krishna, Bransom, Kuehl, Iyyer, Dasigi, Cohan, and Lo}]{krishna_longeval_2023}
Kalpesh Krishna, Erin Bransom, Bailey Kuehl, Mohit Iyyer, Pradeep Dasigi, Arman Cohan, and Kyle Lo. 2023.
\newblock \href {http://arxiv.org/abs/2301.13298} {{LongEval}: {Guidelines} for {Human} {Evaluation} of {Faithfulness} in {Long}-form {Summarization}}.
\newblock \emph{arXiv preprint}.
\newblock ArXiv:2301.13298 [cs].

\bibitem[{Kryściński et~al.(2021)Kryściński, Rajani, Agarwal, Xiong, and Radev}]{kryscinski_booksum_2021}
Wojciech Kryściński, Nazneen Rajani, Divyansh Agarwal, Caiming Xiong, and Dragomir Radev. 2021.
\newblock \href {http://arxiv.org/abs/2105.08209} {{BookSum}: {A} {Collection} of {Datasets} for {Long}-form {Narrative} {Summarization}}.
\newblock \emph{arXiv preprint}.
\newblock ArXiv:2105.08209 [cs].

\bibitem[{Laban et~al.(2021)Laban, Schnabel, Bennett, and Hearst}]{laban_summac_2021}
Philippe Laban, Tobias Schnabel, Paul~N. Bennett, and Marti~A. Hearst. 2021.
\newblock \href {http://arxiv.org/abs/2111.09525} {{SummaC}: {Re}-{Visiting} {NLI}-based {Models} for {Inconsistency} {Detection} in {Summarization}}.
\newblock \emph{arXiv preprint}.
\newblock ArXiv:2111.09525 [cs].

\bibitem[{Lin(2004)}]{lin_rouge_2004}
Chin-Yew Lin. 2004.
\newblock \href {https://aclanthology.org/W04-1013} {{ROUGE}: {A} {Package} for {Automatic} {Evaluation} of {Summaries}}.
\newblock In \emph{Text {Summarization} {Branches} {Out}}, pages 74--81, Barcelona, Spain. Association for Computational Linguistics.

\bibitem[{Liu and Keller(2023)}]{liu_detecting_2023}
Danyang Liu and Frank Keller. 2023.
\newblock \href {http://arxiv.org/abs/2303.17647} {Detecting and {Grounding} {Important} {Characters} in {Visual} {Stories}}.
\newblock \emph{arXiv preprint}.
\newblock ArXiv:2303.17647 [cs].

\bibitem[{Liu et~al.(2020)Liu, Li, Yu, Huang, Liu, Zhao, and Yan}]{liu_character-centric_2020}
Danyang Liu, Juntao Li, Meng-Hsuan Yu, Ziming Huang, Gongshen Liu, Dongyan Zhao, and Rui Yan. 2020.
\newblock \href {https://doi.org/10.1609/aaai.v34i02.5536} {A {Character}-{Centric} {Neural} {Model} for {Automated} {Story} {Generation}}.
\newblock \emph{Proceedings of the AAAI Conference on Artificial Intelligence}, 34(02):1725--1732.
\newblock Number: 02.

\bibitem[{Liu et~al.(2019)Liu, Ott, Goyal, Du, Joshi, Chen, Levy, Lewis, Zettlemoyer, and Stoyanov}]{liu_roberta_2019}
Yinhan Liu, Myle Ott, Naman Goyal, Jingfei Du, Mandar Joshi, Danqi Chen, Omer Levy, Mike Lewis, Luke Zettlemoyer, and Veselin Stoyanov. 2019.
\newblock \href {http://arxiv.org/abs/1907.11692} {{RoBERTa}: {A} {Robustly} {Optimized} {BERT} {Pretraining} {Approach}}.
\newblock \emph{arXiv preprint}.
\newblock ArXiv:1907.11692 [cs].

\bibitem[{Loshchilov and Hutter(2019)}]{loshchilov_decoupled_2019}
Ilya Loshchilov and Frank Hutter. 2019.
\newblock \href {http://arxiv.org/abs/1711.05101} {Decoupled {Weight} {Decay} {Regularization}}.
\newblock \emph{arXiv preprint}.
\newblock ArXiv:1711.05101 [cs, math].

\bibitem[{Maddela et~al.(2022)Maddela, Kulkarni, and Preotiuc-Pietro}]{maddela_entsum_2022}
Mounica Maddela, Mayank Kulkarni, and Daniel Preotiuc-Pietro. 2022.
\newblock \href {https://doi.org/10.18653/v1/2022.acl-long.237} {{EntSUM}: {A} {Data} {Set} for {Entity}-{Centric} {Extractive} {Summarization}}.
\newblock In \emph{Proceedings of the 60th {Annual} {Meeting} of the {Association} for {Computational} {Linguistics} ({Volume} 1: {Long} {Papers})}, pages 3355--3366, Dublin, Ireland. Association for Computational Linguistics.

\bibitem[{Mahon and Lapata(2024)}]{mahon2024modular}
Louis Mahon and Mirella Lapata. 2024.
\newblock \href {https://arxiv.org/abs/2403.03823} {A modular approach for multimodal summarization of tv shows}.
\newblock \emph{Preprint}, arXiv:2403.03823.

\bibitem[{Min et~al.(2023)Min, Krishna, Lyu, Lewis, Yih, Koh, Iyyer, Zettlemoyer, and Hajishirzi}]{min-etal-2023-factscore}
Sewon Min, Kalpesh Krishna, Xinxi Lyu, Mike Lewis, Wen-tau Yih, Pang Koh, Mohit Iyyer, Luke Zettlemoyer, and Hannaneh Hajishirzi. 2023.
\newblock \href {https://doi.org/10.18653/v1/2023.emnlp-main.741} {{FA}ct{S}core: Fine-grained atomic evaluation of factual precision in long form text generation}.
\newblock In \emph{Proceedings of the 2023 Conference on Empirical Methods in Natural Language Processing}, pages 12076--12100, Singapore. Association for Computational Linguistics.

\bibitem[{Nallapati et~al.(2016)Nallapati, Zhou, santos, Gulcehre, and Xiang}]{nallapati_abstractive_2016}
Ramesh Nallapati, Bowen Zhou, Cicero Nogueira~dos santos, Caglar Gulcehre, and Bing Xiang. 2016.
\newblock \href {http://arxiv.org/abs/1602.06023} {Abstractive {Text} {Summarization} {Using} {Sequence}-to-{Sequence} {RNNs} and {Beyond}}.
\newblock \emph{arXiv preprint}.
\newblock ArXiv:1602.06023 [cs] version: 5.

\bibitem[{Narayan et~al.(2018)Narayan, Cohen, and Lapata}]{narayan_dont_2018}
Shashi Narayan, Shay~B. Cohen, and Mirella Lapata. 2018.
\newblock \href {http://arxiv.org/abs/1808.08745} {Don't {Give} {Me} the {Details}, {Just} the {Summary}! {Topic}-{Aware} {Convolutional} {Neural} {Networks} for {Extreme} {Summarization}}.
\newblock \emph{arXiv preprint}.
\newblock ArXiv:1808.08745 [cs] version: 1.

\bibitem[{Narayan et~al.(2022)Narayan, Maynez, Amplayo, Ganchev, Louis, Huot, Das, and Lapata}]{narayan_conditional_2022}
Shashi Narayan, Joshua Maynez, Reinald~Kim Amplayo, Kuzman Ganchev, Annie Louis, Fantine Huot, Dipanjan Das, and Mirella Lapata. 2022.
\newblock \href {http://arxiv.org/abs/2207.00397} {Conditional {Generation} with a {Question}-{Answering} {Blueprint}}.
\newblock \emph{arXiv preprint}.
\newblock ArXiv:2207.00397 [cs].

\bibitem[{Nie et~al.(2020)Nie, Williams, Dinan, Bansal, Weston, and Kiela}]{nie_adversarial_2020}
Yixin Nie, Adina Williams, Emily Dinan, Mohit Bansal, Jason Weston, and Douwe Kiela. 2020.
\newblock \href {https://doi.org/10.18653/v1/2020.acl-main.441} {Adversarial {NLI}: {A} {New} {Benchmark} for {Natural} {Language} {Understanding}}.
\newblock In \emph{Proceedings of the 58th {Annual} {Meeting} of the {Association} for {Computational} {Linguistics}}, pages 4885--4901, Online. Association for Computational Linguistics.

\bibitem[{Papalampidi et~al.(2019)Papalampidi, Keller, and Lapata}]{papalampidi_movie_2019}
Pinelopi Papalampidi, Frank Keller, and Mirella Lapata. 2019.
\newblock \href {https://doi.org/10.18653/v1/D19-1180} {Movie {Plot} {Analysis} via {Turning} {Point} {Identification}}.
\newblock In \emph{Proceedings of the 2019 {Conference} on {Empirical} {Methods} in {Natural} {Language} {Processing} and the 9th {International} {Joint} {Conference} on {Natural} {Language} {Processing} ({EMNLP}-{IJCNLP})}, pages 1707--1717, Hong Kong, China. Association for Computational Linguistics.

\bibitem[{Phelan(1989)}]{phelan1989reading}
James Phelan. 1989.
\newblock \emph{Reading people, reading plots: Character, progression, and the interpretation of narrative}.
\newblock University of Chicago Press.

\bibitem[{Raffel et~al.(2020)Raffel, Shazeer, Roberts, Lee, Narang, Matena, Zhou, Li, and Liu}]{raffel_exploring_2020}
Colin Raffel, Noam Shazeer, Adam Roberts, Katherine Lee, Sharan Narang, Michael Matena, Yanqi Zhou, Wei Li, and Peter~J. Liu. 2020.
\newblock \href {http://arxiv.org/abs/1910.10683} {Exploring the {Limits} of {Transfer} {Learning} with a {Unified} {Text}-to-{Text} {Transformer}}.
\newblock \emph{arXiv preprint}.
\newblock ArXiv:1910.10683 [cs, stat].

\bibitem[{Rajpurkar et~al.(2016)Rajpurkar, Zhang, Lopyrev, and Liang}]{rajpurkar_squad_2016}
Pranav Rajpurkar, Jian Zhang, Konstantin Lopyrev, and Percy Liang. 2016.
\newblock \href {https://doi.org/10.18653/v1/D16-1264} {{SQuAD}: 100,000+ {Questions} for {Machine} {Comprehension} of {Text}}.
\newblock In \emph{Proceedings of the 2016 {Conference} on {Empirical} {Methods} in {Natural} {Language} {Processing}}, pages 2383--2392, Austin, Texas. Association for Computational Linguistics.

\bibitem[{Robertson et~al.(1995)Robertson, Walker, Jones, Hancock-Beaulieu, and Gatford}]{robertson1995okapi}
Stephen Robertson, S.~Walker, S.~Jones, M.~M. Hancock-Beaulieu, and M.~Gatford. 1995.
\newblock \href {https://www.microsoft.com/en-us/research/publication/okapi-at-trec-3/} {Okapi at trec-3}.
\newblock In \emph{Overview of the Third Text REtrieval Conference (TREC-3)}, pages 109--126. Gaithersburg, MD: NIST.

\bibitem[{Sang et~al.(2022)Sang, Mou, Yu, Wang, Li, and Stanton}]{sang_mbti_2022}
Yisi Sang, Xiangyang Mou, Mo~Yu, Dakuo Wang, Jing Li, and Jeffrey Stanton. 2022.
\newblock \href {http://arxiv.org/abs/2210.10994} {{MBTI} {Personality} {Prediction} for {Fictional} {Characters} {Using} {Movie} {Scripts}}.
\newblock \emph{arXiv preprint}.
\newblock ArXiv:2210.10994 [cs].

\bibitem[{Song et~al.(2024)Song, Kim, and Iyyer}]{song_veriscore_2024}
Yixiao Song, Yekyung Kim, and Mohit Iyyer. 2024.
\newblock \href {http://arxiv.org/abs/2406.19276} {{VERISCORE}: {Evaluating} the factuality of verifiable claims in long-form text generation}.
\newblock \emph{arXiv preprint}.
\newblock ArXiv:2406.19276 [cs].

\bibitem[{Vaswani et~al.(2017)Vaswani, Shazeer, Parmar, Uszkoreit, Jones, Gomez, Kaiser, and Polosukhin}]{vaswani_attention_2017}
Ashish Vaswani, Noam Shazeer, Niki Parmar, Jakob Uszkoreit, Llion Jones, Aidan~N. Gomez, Lukasz Kaiser, and Illia Polosukhin. 2017.
\newblock \href {http://arxiv.org/abs/1706.03762} {Attention {Is} {All} {You} {Need}}.
\newblock \emph{arXiv preprint}.
\newblock ArXiv:1706.03762 [cs].

\bibitem[{Weiland(2016)}]{Weiland:2016}
K.M. Weiland. 2016.
\newblock \emph{Creating Character Arcs}.
\newblock PenForASword Publishing.

\bibitem[{Wu et~al.(2021)Wu, Ouyang, Ziegler, Stiennon, Lowe, Leike, and Christiano}]{wu_recursively_2021}
Jeff Wu, Long Ouyang, Daniel~M. Ziegler, Nisan Stiennon, Ryan Lowe, Jan Leike, and Paul Christiano. 2021.
\newblock \href {http://arxiv.org/abs/2109.10862} {Recursively {Summarizing} {Books} with {Human} {Feedback}}.
\newblock \emph{arXiv preprint}.
\newblock ArXiv:2109.10862 [cs].

\bibitem[{Xu et~al.(2023)Xu, Ping, Wu, McAfee, Zhu, Liu, Subramanian, Bakhturina, Shoeybi, and Catanzaro}]{xu_retrieval_2023}
Peng Xu, Wei Ping, Xianchao Wu, Lawrence McAfee, Chen Zhu, Zihan Liu, Sandeep Subramanian, Evelina Bakhturina, Mohammad Shoeybi, and Bryan Catanzaro. 2023.
\newblock \href {http://arxiv.org/abs/2310.03025} {Retrieval meets {Long} {Context} {Large} {Language} {Models}}.
\newblock \emph{arXiv preprint}.
\newblock ArXiv:2310.03025 [cs].

\bibitem[{Zaheer et~al.(2021)Zaheer, Guruganesh, Dubey, Ainslie, Alberti, Ontanon, Pham, Ravula, Wang, Yang, and Ahmed}]{zaheer_big_2021}
Manzil Zaheer, Guru Guruganesh, Avinava Dubey, Joshua Ainslie, Chris Alberti, Santiago Ontanon, Philip Pham, Anirudh Ravula, Qifan Wang, Li~Yang, and Amr Ahmed. 2021.
\newblock \href {http://arxiv.org/abs/2007.14062} {Big {Bird}: {Transformers} for {Longer} {Sequences}}.
\newblock \emph{arXiv preprint}.
\newblock ArXiv:2007.14062 [cs, stat] version: 2.

\bibitem[{Zhang et~al.(2020)Zhang, Kishore, Wu, Weinberger, and Artzi}]{zhang_bertscore_2020}
Tianyi Zhang, Varsha Kishore, Felix Wu, Kilian~Q. Weinberger, and Yoav Artzi. 2020.
\newblock \href {http://arxiv.org/abs/1904.09675} {{BERTScore}: {Evaluating} {Text} {Generation} with {BERT}}.
\newblock \emph{arXiv preprint}.
\newblock ArXiv:1904.09675 [cs].

\bibitem[{Zhang et~al.(2019)Zhang, Cheung, and Oren}]{zhang_generating_2019}
Weiwei Zhang, Jackie Chi~Kit Cheung, and Joel Oren. 2019.
\newblock \href {https://doi.org/10.1609/aaai.v33i01.33017476} {Generating {Character} {Descriptions} for {Automatic} {Summarization} of {Fiction}}.
\newblock \emph{Proceedings of the AAAI Conference on Artificial Intelligence}, 33(01):7476--7483.
\newblock Number: 01.

\bibitem[{Zhao et~al.(2022)Zhao, Brahman, Song, Yao, Yu, and Chaturvedi}]{zhao_narrasum_2022}
Chao Zhao, Faeze Brahman, Kaiqiang Song, Wenlin Yao, Dian Yu, and Snigdha Chaturvedi. 2022.
\newblock \href {http://arxiv.org/abs/2212.01476} {{NarraSum}: {A} {Large}-{Scale} {Dataset} for {Abstractive} {Narrative} {Summarization}}.
\newblock \emph{arXiv preprint}.
\newblock ArXiv:2212.01476 [cs].

\bibitem[{Zheng et~al.(2023)Zheng, Chiang, Sheng, Zhuang, Wu, Zhuang, Lin, Li, Li, Xing, Zhang, Gonzalez, and Stoica}]{zheng_judging_2023}
Lianmin Zheng, Wei-Lin Chiang, Ying Sheng, Siyuan Zhuang, Zhanghao Wu, Yonghao Zhuang, Zi~Lin, Zhuohan Li, Dacheng Li, Eric~P. Xing, Hao Zhang, Joseph~E. Gonzalez, and Ion Stoica. 2023.
\newblock \href {https://doi.org/10.48550/arXiv.2306.05685} {Judging {LLM}-as-a-{Judge} with {MT}-{Bench} and {Chatbot} {Arena}}.
\newblock \emph{arXiv preprint}.
\newblock ArXiv:2306.05685 [cs].

\end{thebibliography}

\clearpage
\appendix

\section{\textsc{BookWorm} Descriptions and Analyses} \label{sec:data_examples}
We identify the subset of \textsc{BookWorm} that includes characters featured in both the description and analysis tasks. We then compare the two tasks by calculating the percentage of novel n-grams in descriptions compared to analyses. We present these statistics in Table~\ref{tab:comparison_descr_analysis}.

\begin{table}[ht]
\centering

\begin{tabular}{lccc}
\toprule
 & \multicolumn{3}{c}{\textbf{Novel n-grams \%}} \\
& \multicolumn{1}{c}{Unigrams} & Bigrams & Trigrams \\
\midrule
\textsc{BookWorm}  & 49.60 & 83.81	& 95.96	 \\
\bottomrule
\end{tabular} 

\caption{Percentage of novel n-grams (unigrams, bigrams, trigrams) in character descriptions compared to analyses.}
\label{tab:comparison_descr_analysis}
\end{table}

We show examples of gold-standard descriptions and analyses in Tables~\ref{tab:examples}--\ref{tab:examples4}.

\section{Implementation Details} \label{sec:train_details}
We fully fine-tune LongT5-base~(250M), which takes approximately 10 GPU hours for the description task and 4 GPU hours for the analysis task. For Llama-3-8B, we use parameter-efficient fine-tuning with LoRA, which takes roughly 8 and 3 GPU hours for the description and analysis tasks, respectively. We fine-tune our models using the AdamW~\cite{loshchilov_decoupled_2019} optimizer. We use batch size equal to 1 and a gradient accumulation step equal to 4. For LongT5, we use a constant learning rate of 1e-4, while for the Llama-3 model, we use a learning rate of 2e-5 with linear decay. For Low-Rank fine-tuning, we use a rank of 8 and an alpha of 16.

For evaluation, we use a publicly available Rouge-score implementation\footnote{\url{https://github.com/google-research/google-research/tree/master/rouge}} and for BERTScore we use DeBERTa-Xlarge~\cite{he_deberta_2021}. We calculate entity mention recall using a named-entity recognition module from Spacy\footnote{\url{https://spacy.io/}}.

We present the prompt used for extracting facts, which is adapted from the VeriScore metric~\cite{song_veriscore_2024} to suit our specific tasks in Table~\ref{tab:fact_extraction}. The prompt used to verify whether a fact is supported is shown in Table~\ref{tab:verify_fact}.
We provide the prompt for classifying facts in Table~\ref{tab:prompt_for_classification} and the instructions given to the human annotators in Table~\ref{tab:instructions}.

For our experiments, we use the following prompts:
\begin{lstlisting}[linewidth=\columnwidth, breaklines=true, language={}, captionpos=b, title=Character Description Prompt ]
Describe character: {character_name} given the following context. Context: [..]
\end{lstlisting}

\begin{lstlisting}[linewidth=\columnwidth, breaklines=true, language={}, captionpos=b, title=Character Analysis Prompt]
Analyse in-depth character: {character_name} given the following context. Context: [..]
\end{lstlisting}

\begin{lstlisting}[linewidth=\columnwidth, breaklines=true, language={}, captionpos=b, title=Character Description Prompt with character names]
The following story has these characters: {list_of_character} describe character: {character_name} given the following context. Context: [..]
\end{lstlisting}


\begin{lstlisting}[linewidth=\columnwidth, breaklines=true, language={}, captionpos=b, title=Joint Character Description Prompt]
Describe the following characters: {list_of_character} given the following context and return your output as in the following examples \n\n {example_1} \n\n {example_2} \n\n {example_3}. Context: [..]
\end{lstlisting}

\section{Additional Experiments} \label{sec:additional_experiments}

We ran a ``no-context'' experiment where only the initial prompt was provided, without including any part of the story. This is a way of testing how much of the story has been memorized by the pre-trained LLMs we use for our work. This provides an indication of the degree of data contamination that is present.

In another experiment, we used a summary of the story as input instead of the book text. This is the setup that has been used by other work on character description generation, most notably by \citet{brahman_let_2021}. We ran these experiments using the Llama-3-8B model in a zero-shot way.

We also evaluate the performance of the closed-source model, GPT-4o-mini, in two different ways, first processing the lead 128k tokens of the input story and second extracting context with the coreference-based approach. 

The results are presented in Table~\ref{tab:additional_results} and compared against the performance of the coreference-based approach. For the description task, the no-context setting performs competitively, though it falls behind the coreference-based approach. The summary-based model outperformed the no-context approach, being slightly better in Rouge-L than the coref-based model and has a clear advantage in terms of PRISMA score. However, it underperforms compared to the retrieval-augmented method in entity mention recall and question answering metric. GPT-4o-mini with 128k input context achieves the highest scores in entities recall and question answering F1, although it has a slightly lower Rouge-L than the Llama experiments. GPT-4o-mini with retrieved-context is worse than the Lead-128k experiment but surpasses the Llama experiments in entities recall and QA-F1 while being weaker in Rouge-L.

For the analysis task, the no-context model outperforms the retrieval-augmented approach across all metrics. This indicates a high degree of contamination for this task, i.e., the pretrained models have seen the texts in the \textsc{BookWorm} testset during pretraining. The summary-based approach achieves increased scores surpassing both the no-context and retrieval-augmented models. This improvement can likely be attributed to the fact that the summaries were sourced from the same websites used to scrape the description and analysis data, leading to high lexical overlap and thus providing an unintended advantage. Similar to the description task, GPT-4o-mini obtains the highest results in entities mention recall and QA-F1 metrics but falls behind in terms of Rouge-L. The coreference-based GPT-4o-mini model is only marginally worse than the full context experiment.

\section{Example Output} \label{sec:qa_based_eval}

We show examples of model output for the character description task in Tables~\ref{tab:eval_examples1}--\ref{tab:fact_eval_example1}, for the analysis task in Tables~\ref{tab:eval_examples4}--\ref{tab:fact_eval_example2} and for joint character description in Table~\ref{tab:joint_example}. 

\onecolumn

\begin{table*}[ht]

\centering
\begin{tabular}{@{}p{0.98\textwidth}@{}}
\toprule
\textbf{Character:}  Alice \quad  \textbf{Book:} Alice in Wonderland  by Lewis Carroll\\
\midrule
\textbf{Source:} GradeSaver \\
\textbf{Description:} The heroine of the story. Her adventures begin with her fateful jump down the rabbit hole, and the tale is an extended metaphor for the challenges she will face as she grows into an adult. She possesses unusual composure for a child, and she seems bright but makes many charming mistakes. She grows more confident as the book progresses.
\\
\midrule
\textbf{Source:} SparkNotes \\
\textbf{Analysis:}
Alice is a sensible prepubescent girl from a wealthy English family who finds herself in a strange world ruled by imagination and fantasy. Alice feels comfortable with her identity and has a strong sense that her environment is comprised of clear, logical, and consistent rules and features. Alice’s familiarity with the world has led one critic to describe her as a “disembodied intellect.” Alice displays great curiosity and attempts to fit her diverse experiences into a clear understanding of the world.
Alice approaches Wonderland as an anthropologist, but maintains a strong sense of noblesse oblige that comes with her class status. She has confidence in her social position, education, and the Victorian virtue of good manners. Alice has a feeling of entitlement, particularly when comparing herself to Mabel, whom she declares has a “poky little house,” and no toys. Additionally, she flaunts her limited information base with anyone who will listen and becomes increasingly obsessed with the importance of good manners as she deals with the rude creatures of Wonderland. Alice maintains a superior attitude and behaves with solicitous indulgence toward those she believes are less privileged.
The tension of Alice’s Adventures in Wonderland emerges when Alice’s fixed perspective of the world comes into contact with the mad, illogical world of Wonderland. Alice’s fixed sense of order clashes with the madness she finds in Wonderland. The White Rabbit challenges her perceptions of class when he mistakes her for a servant, while the Mad Hatter, March Hare, and Pigeon challenge Alice’s notions of urbane intelligence with an unfamiliar logic that only makes sense within the context of Wonderland. Most significantly, Wonderland challenges her perceptions of good manners by constantly assaulting her with dismissive rudeness. Alice’s fundamental beliefs face challenges at every turn, and as a result Alice suffers an identity crisis. She persists in her way of life as she perceives her sense of order collapsing all around her. Alice must choose between retaining her notions of order and assimilating into Wonderland’s nonsensical rules.\\

\bottomrule
\end{tabular}
\caption{Example of gold-standard description and analysis in \textsc{BookWorm}.}
\label{tab:examples}
\end{table*}

\begin{table*}[ht]

\centering
\begin{tabular}{@{}p{0.98\textwidth}@{}}
\toprule
\textbf{Character:}  Katherine \quad  \textbf{Book:} The Taming of the Shrew by William Shakespeare \\
\midrule
\textbf{Source:} SparkNotes \\
\textbf{Description:} The “shrew” of the play’s title, Katherine, or Kate, is the daughter of Baptista Minola, with whom she lives in Padua. She is sharp-tongued, quick-tempered, and prone to violence, particularly against anyone who tries to marry her. Her hostility toward suitors particularly distresses her father. But her anger and rudeness disguise her deep-seated sense of insecurity and her jealousy toward her sister, Bianca. She does not resist her suitor Petruchio forever, though, and she eventually subjugates herself to him, despite her previous repudiation of marriage.
\\
\midrule
\textbf{Source:} SparkNotes \\
\textbf{Analysis:} Widely reputed throughout Padua to be a shrew, Katherine is foul-tempered and sharp-tongued at the start of the play. She constantly insults and degrades the men around her, and she is prone to wild displays of anger, during which she may physically attack whomever enrages her. Though most of the play’s characters simply believe Katherine to be inherently ill-tempered, it is certainly plausible to think that her unpleasant behavior stems from unhappiness. She may act like a shrew because she is miserable and desperate. There are many possible sources of Katherine’s unhappiness: she expresses jealousy about her father’s treatment of her sister, but her anxiety may also stem from feelings about her own undesirability, the fear that she may never win a husband, her loathing of the way men treat her, and so on. In short, Katherine feels out of place in her society. Due to her intelligence and independence, she is unwilling to play the role of the maiden daughter. She clearly abhors society’s expectations that she obey her father and show grace and courtesy toward her suitors. At the same time, however, Katherine must see that given the rigidity of her social situation, her only hope to find a secure and happy place in the world lies in finding a husband. These inherently conflicting impulses may lead to her misery and poor temper. A vicious circle ensues: the angrier she becomes, the less likely it seems she will be able to adapt to her prescribed social role; the more alienated she becomes socially, the more her anger grows.
Despite the humiliations and deprivations that Petruchio adds to her life, it is easy to understand why Katherine might succumb to marry a man like him. In their first conversation, Petruchio establishes that he is Katherine’s intellectual and verbal equal, making him, on some level, an exciting change from the easily dominated men who normally surround her. Petruchio’s forcible treatment of Katherine is in every way designed to show her that she has no real choice but to adapt to her social role as a wife. This adaptation must be attractive to Katherine on some level, since even if she dislikes the role of wife, playing it at least means she can command respect and consideration from others rather than suffer the universal revulsion she receives as a shrew. Having a social role, even if it is not ideal, must be less painful than continually rejecting any social role at all. Thus, Katherine’s eventual compliance with Petruchio’s self-serving “training” appears more rational than it might have seemed at first: by the end of the play, she has gained a position and even an authoritative voice that she previously had been denied.
\\
\bottomrule
\end{tabular}
\caption{Example of gold-standard description and analysis in \textsc{BookWorm}.}
\label{tab:examples2}
\end{table*}

\begin{table*}[ht]

\centering
\begin{tabular}{@{}p{0.98\textwidth}@{}}
\toprule
\textbf{Character:} Prospero \quad  \textbf{Book:} The Tempest by William Shakespeare \\
\midrule
\textbf{Source:} GradeSaver \\
\textbf{Description:} The rightful Duke of Milan, though his kingdom and title were usurped by his brother Antonio. Prospero was able to survive a plot on his life, and he and his daughter Miranda were set aboard a wrecked craft, but managed to land safely on the island. Prospero is able to gain control of the spirits of the island, and uses his vast knowledge and control over the spirits to direct acts of magic as he pleases. He is ruler of the island, after taking control of it from its rightful heir, Caliban, and he makes sure that Alonso's ship wrecks on the island, so he can get his revenge on his brothers for their wrongdoing.
\\
\midrule
\textbf{Source:} Cliffnotes \\
\textbf{Analysis:} Prospero is the rightful duke of Milan. Twelve years earlier, he found refuge on this island after his younger brother, Antonio, seized Prospero's title and property. Prospero functions as a god on the island, manipulating everyone within his reach. He is helpless against his enemies until they appear on a ship nearby; but when they are close enough, he can use his magic to create a storm and bring them under his control.
Prospero's magic is the white magic of nature, not the black magic of evil men. This former duke of Milan is a complex personality. Although he refuses to free Ariel and enslaves Caliban, Prospero is really a beneficent ruler, never intending to injure even his enemies. Early in the play, Prospero appears callous and cruel, especially in his treatment of Ariel and Caliban. He is also autocratic in his treatment of Ferdinand, but Prospero realizes that Ferdinand and Miranda will value one another more if there are a few impediments to their courtship.
Prospero's humanity is clearly obvious in his treatment of Antonio, whom he calls traitor but whom he declines to treat as a traitor. Another example of Prospero's goodness is when he stops Alonso from apologizing to Miranda, telling him that there is no need for more amends. By the play's conclusion, it is clear that Prospero is just and fair, in addition to intelligent.\\
\bottomrule
\end{tabular}
\caption{Example of gold-standard description and analysis in \textsc{BookWorm}.}
\label{tab:examples3}
\end{table*}

\begin{table*}[ht]

\centering
\begin{tabular}{@{}p{0.98\textwidth}@{}}
\toprule
\textbf{Character:}  Stephen Dedalus \quad  \textbf{Book:} Ulysses by James Joyce \\
\midrule
\textbf{Source:} GradeSaver \\
\textbf{Description:} An aspiring poet in his early twenties. Stephen is intelligent and extremely well-read, and he likes music. He seems to exist more for himself, in a cerebral way, than as a member of a community or even the group of medical students that he associates with. Stephen was extremely religious as a child, but now he struggles with issues of faith and doubt in the wake of his mother’s death,which occurred less than a year ago.
\\
\midrule
\textbf{Source:} Cliffnotes \\
\textbf{Analysis:} In A Portrait of the Artist as a Young Man, Stephen was treated with both irony and sympathy. Joyce admired his young protagonist's battle against orthodoxy, but he also found Stephen's intolerant cynicism a bit pompous. In Book Five of A Portrait, Stephen became a mock Christ figure, preaching his gospel of aesthetics to bored and sometimes gibing apostles. In Ulysses, Stephen is a more human figure than he appeared to be at the end of the earlier novel. He has returned from Paris, his destination at the end of A Portrait, having been summoned home by word of his mother' s incipient death from cancer; now he finds himself emotionally drowning as surely as his mother literally drowned in her own green bile. In Ulysses, he sees himself as an Icarus-like figure, one who flew too high and burned his wings in the sun; as "Daedalus," he parallels himself with the archetypal flying ace.

In Ulysses, Stephen is beset with many problems, some of them stemming from his emotional distance from those around him, whom he cannot accept. Although he lives in the Martello Tower with Haines, the Oxonian, and with Buck Mulligan, a Dublin medical student, he knows that he cannot remain in this habitat: Haines has bizarre nightmares that keep Stephen awake, and Mulligan, with his coarse and brutal treatment of Stephen, has "usurped" Stephen's place in the Tower. At the end of "Telemachus," he meekly surrenders the Tower's key to Mulligan and begins to walk his own path. Compared to the physical Mulligan, Stephen feels himself to be inept and weak. Stephen is afraid of water (symbolically, baptism), while Mulligan plunges into life. In many ways, Stephen is physically withdrawn, fearing dogs and thunder, while Mulligan once saved a man from drowning. The facile Mulligan can handle the visiting milk woman in "Telemachus," although he looks down upon her, while Stephen sits brooding upon the lost past of Ireland. Stephen's estrangement is also seen in his teaching at Mr. Deasy's school, where he does not seem to care, really, that his students are inattentive and obstreperous.

Stephen's sense of abstraction, of distance, forces him to turn inward for answers, and, it is through Joyce's presentation of Stephen's vexed psyche and soul, especially in "Proteus," that we see his young man's bewilderment over the changing, "protean," nature of reality. Divested of his former stringent religious beliefs, wishing to become a famous writer though sometimes doubting his ability to do so, Stephen, in "Proteus," is searching for his origins. He imagines that the two old women that he sees on the beach are midwives; he projects an image of navel cords linking all humanity and ending with Eve, "belly without blemish." He wonders who his real father is: Simon, whose part in Stephen's conception was physiological; or God Himself, Who may have planned the event from all eternity.

Stephen's ruminations lead him to feel a great sense of guilt, which is augmented by his tender conscience, one that focuses upon blemishes and ignores virtues. Stephen feels guilty for many things: he refused to pray at his dying mother's bedside; he smokes Haines's tobacco, yet he treats him with disdain; he borrowed a pound from the theosophist George Russell (A. E.) and spent it on a prostitute; as the eldest Dedalus child, he abandoned his starving sisters to a poverty which was worsened by an alcoholic father who spends his time in bars while the family barely survives; he led a false existence when he was a youth, pretending so well that he was deeply pious that he was singled out for training in the priesthood, yet all the time, he was thinking of naked women.
 [truncated]
\\
\bottomrule
\end{tabular}
\caption{Example of gold-standard description and analysis in \textsc{BookWorm}.}
\label{tab:examples4}
\end{table*}

\begin{table*}[ht]

\centering
\begin{tabular}{@{}p{0.98\textwidth}@{}}
\toprule

You are trying to verify how factual a piece of text is. To do so, you need to break down a sentence and extract as many fine-grained facts mentioned in the sentence as possible. 
\\ \\
Extract fine-grained facts from the sentence marked between <SOS> and <EOS>. You should focus on the named entities and numbers in the sentence and extract relevant information from the sentence. Other sentences are only context for you to recover pronouns, definite phrases (e.g., "the victims" or "the pope"), and so on. Each fact should be understandable on its own and require no additional context. This means that all entities must be referred to by name but not pronoun. Use the name of entities rather than definite noun phrases (e.g., 'the teacher') whenever possible. If a definite noun phrase is used, be sure to add modifiers (e.g., a embedded clause, a prepositional phrase, etc.). Each fact must be situated within relevant temporal and location whenever needed. Keep each fact to one sentence with zero or at most one embedded clause. You do not need to justify what you extract.
\\ \\
Here are some examples:
\{examples\}
\\ \\
Text: \{snippet\} \\
Sentence to be focused on: \{sentence\} \\
Facts:
\\ 
\bottomrule
\end{tabular}
\caption{Prompt used to extract facts for a given sentence. This prompt is adjusted based on the VeriScore metric~\cite{song_veriscore_2024}.}
\label{tab:fact_extraction}
\end{table*}

\begin{table*}[ht]

\centering
\begin{tabular}{@{}p{0.98\textwidth}@{}}
\toprule

Classify the following fact as either true or false based on the provided context.
\\ \\
Examples: \{examples\}

Your task \\ \\
Context: \{context\} \\ \\
Fact:\{fact\} True or False? \\ \\
Output: \\
\bottomrule 
\end{tabular}
\caption{Prompt to verify a fact given the context. The prompt is adapted from the FactScore metric~\cite{min-etal-2023-factscore}.}
\label{tab:verify_fact}
\end{table*}

\begin{table*}[ht]

\centering
\begin{tabular}{@{}p{0.98\textwidth}@{}}
\toprule

Can you classify the following fact of a given character description/analysis in one of following categories: Role, Relationship, Event, Personality, Mental State or Other Fact.
\\ \\ 
Role: defines what part the character plays in the story, narrator, major/minor character. \\
Relationship: describes the connections the character has with others, such as friendships or family ties, which influence their actions and decisions. \\
Personality: Personality encompasses the character's behavior, traits, and attributes, shaping how they think, act, and interact with others. \\
Events: key actions and decisions the character is involved in throughout the story, driving the plot forward. \\
Mental State: state of mind of a character including cognition, beliefs, intentions and emotions. Mental state can fluctuate (in contrast with personality which is something more permanent). Common verbs to express mental state would be 'think', 'believe', 'know'. \\
Other Fact: any other fact that doesn't belong to the above categories.  Return as output just a single or two words. \\ 
\\ 
Character Description/Analysis: \{description\}
\\ \\ 
Fact: \{fact\}
\\ 
Output: \\
\bottomrule
\end{tabular}
\caption{Prompt used to classify the extracted facts in one of the subcategories.}
\label{tab:prompt_for_classification}
\end{table*}

\begin{table*}[ht]

\centering
\begin{tabular}{@{}p{0.98\textwidth}@{}}
\toprule
\textbf{Facts Annotation} \\ \\

\textbf{Instructions}
\\
You will be given character descriptions/analyses along with their corresponding extracted facts. Your task is to classify each one of the facts in one of the following categories: Role, Relationship, Event, Personality, Mental State, Other Fact. \\

\textbf{Role}: defines what part the character plays in the story, narrator, major/minor character. \\
\textbf{Relationship}: connections the character has with others, such as friendships or family ties. \\
\textbf{Personality}: character's behavior, traits, and attributes. \\
\textbf{Events}: actions and decisions the character is involved in throughout the story. \\
\textbf{Mental State}: state of mind of a character including cognition, beliefs, intentions and emotions. Mental state can fluctuate (in contrast with personality which is more permanent). Common verbs to express mental state would be "think", "believe", "know". \\
\textbf{Other Fact}: any other fact that doesn't belong to the above categories. \\ \\

In facts where two categories are relevant try to choose the one which is the best fit. Also, if none of the categories corresponds to a fact please choose the category Other Fact. \\ \\ 

\textbf{Examples}
To further explain the annotation task, bellow, you will find an example character description along with the corresponding extracted facts and their correct classifications. A set of examples is also provided to demonstrate edge cases.
\\ \\
You will classify 7 character descriptions with 100 facts in total and 2 character analyses with 100 facts in total.\\

\bottomrule
\end{tabular}
\caption{Instructions provided to human annotators for the classification of facts type.}
\label{tab:instructions}
\end{table*}

\begin{table*}[ht]
\centering
\resizebox{\textwidth}{!}{%
\begin{tabular}{@{}lcccc|cccc}
\toprule
\multirow{3}{*}{\textbf{Model}} & \multicolumn{4}{c}{\textbf{Description}} & \multicolumn{4}{c}{\textbf{Analysis}} \\
& \multicolumn{1}{c}{R-L} & EntMent & QA-F1 & \multicolumn{1}{c}{PRISMA} & \multicolumn{1}{c}{R-L} & EntMent & QA-F1 & PRISMA \\
\midrule
Llama (no-context) & 18.32 & 31.39 & 15.12 & 51.31 & 16.97 & 28.78 & 15.10 & 52.30 \\
\hspace{0.3cm} + summary & \textbf{18.76} & 26.55 & 15.28 & \textbf{63.24}  & \textbf{17.03}  & 29.05 & 15.05 & \textbf{55.89} \\
\hspace{0.3cm} + coref & 18.55 & 32.27 & 16.43 & 52.59 & 16.60 & 22.80 & 14.63 &  52.04  \\
GPT-4o-mini &  18.15 & \textbf{37.01} & \textbf{18.66} & --- & 16.19 & \textbf{31.02} & \textbf{16.91} & --- \\
\hspace{0.3cm} + coref & 17.30 & 34.51 & 18.12 & --- & 16.15  & 30.72 & 16.16 & --- \\
\bottomrule
\end{tabular} 
}
\caption{Zero-shot performance of Llama-3-8B model with no context as input, using the story's summary and a coreference model to extract context. Performance of the closed-source GPT-4o-mini model using the lead 124k input context and the retrieved context from the coreference model.
We report  Rouge-L, entity mention recall (EntMent), question answering F1 and PRISMA-F1 score. We do not compute PRISMA scores for the GPT-4o-mini experiments, as prior work has shown that LLM judges are biased towards their own predictions~\cite{zheng_judging_2023}. Best model per metric is boldfaced.}\label{tab:additional_results}

\end{table*}

\newpage
\clearpage

\begin{table*}[ht]
\centering
\begin{tabular}{@{}p{0.5\textwidth} p{0.22\textwidth} p{0.2\textwidth}@{}}
\toprule
\textbf{Character:}  Admetos \quad  \textbf{Book:} Alcestis  by Euripides\\
\midrule
\multicolumn{3}{@{}p{\textwidth}@{}}{\textbf{Reference Description:} Admetos is the King of Thessaly, a small kingdom in northern Greece. He is married to Alcestis. Before the play begins, Admetos is doomed by the Fates to an early death, but his friend and guest, Apollo, tricks the Fates into sparing him. However, someone must die in his place, and Alcestis volunteers. Admetos is plunged into despair at her death and vows never to remarry or even to enjoy his life ever again. Admetos is portrayed as being genuinely hospitable, generous, and a good friend; at the same time, he is also used to getting his way and has never suffered much before, so he can display an entitled, selfish attitude. He misleads his friend Herakles regarding Alcestis’s death, but Herakles loyally fights Death and brings Alcestis back from the dead to repay his friend’s hospitality. In the process, with the help of the chorus and chorus leader, Admetos realizes that he cannot spend his life dodging death, that love brings suffering, and that he can only know happiness once he accepts mortality. It’s only after he has learned these lessons that he and Alcestis are reunited, prompting Admetos to declare a feast and reform his life.} \\
\midrule
\multicolumn{3}{@{}p{\textwidth}@{}}{\textbf{Llama-3 (zero-shot),  Hierarchical:} King Admetos is a complex character in the play "Alcestis" by Euripides. He is the king of Pherae in Thessaly and the husband of Alcestis, who has died in his place after being granted a special privilege by Apollo. Admetos is a gallant prince who has won the affection of great guests such as Apollo and Heracles, but he is also shown to be selfish and unwilling to accept the reality of his wife's death. } \\
\midrule
\textbf{QA pairs:} &  &  \\
Who is the King of Thessaly? 	& Prediction: \textcolor{blue}{Admetos} &  Answer: Admetos \\
Who tricks the Fates into sparing Admetos from an early death? &	 Prediction: \textcolor{blue}{Apollo} 	& Answer: Apollo \\
Who volunteers to die in Admetos's place? 	& Prediction: \textcolor{blue}{Alcestis} 	& Answer: Alcestis \\
Who brings Alcestis back from the dead to repay Admetos's hospitality? &	 Prediction: \textcolor{Red}{Apollo} 	& Answer: Herakles \\
\midrule
\multicolumn{3}{@{}p{\textwidth}@{}}{\textbf{Llama-3 - coref:} King Admetos is a character who is described by the Third Elder as being the bravest and truest of husbands. He is a king who has been saved from death by the sacrifice of his wife, Alcestis. He is a devoted husband and father who is grateful for the gift of life that his wife has given him. He is also a king who is willing to make sacrifices for his family, as he has given up his life to save his son. } \\
\midrule
\textbf{QA pairs:} &  & \\
Who is the King of Thessaly?  & 	 Prediction: \textcolor{blue}{Admetos}  & 	 Answer: Admetos \\
Who tricks the Fates into sparing Admetos from an early death? & 	 Prediction: \textcolor{Red}{Alcestis}  & 	 Answer: Apollo \\
Who volunteers to die in Admetos's place? & 	 Prediction: \textcolor{blue}{Alcestis} & 	 Answer: Alcestis \\
Who brings Alcestis back from the dead to repay Admetos's hospitality? & 	 Prediction: \textcolor{Red}{Third Elder} &  Answer: Herakles \\
\bottomrule
\end{tabular}
\caption{Examples of generated descriptions and questions with \textcolor{blue}{correct} and  \textcolor{red}{wrong} answers based on them.}
\label{tab:eval_examples1}
\end{table*}

\begin{table*}[ht]
\centering
\begin{tabular}{@{}p{0.4\textwidth} p{0.3\textwidth} p{0.25\textwidth}@{}}
\toprule
\multicolumn{3}{@{}l}{\textbf{Character:} Mary \quad \textbf{Book:} Clotel; or, The President's Daughter by William Wells Brown} \\ 
\midrule
\multicolumn{3}{@{}p{\textwidth}@{}}{\textbf{Reference Description:} Mary is the daughter of Clotel and remains behind once her mother has been sold off as a result of Horatio marrying Gertrude. Left behind to fill the role of a household slave she receives the full brunt of the white Mrs. Green’s residual jealousy toward her mother. The lousy treatment that Mary receives as the Greens’ servant serves to inspire her to take the chance to help a mulatto servant in the home named George escape. As a result of this action, Mary is sold at the New Orleans slave market but is rescued from sharing the same fate as his mother and grandmother by a Frenchman named Devenant with whom she escapes to Europe and marries. After the death of Devenant, George, and Mary cross paths again and wind up marrying to bring the novel to the closest thing to a happy ending that is probably possible.
} \\
\midrule
\multicolumn{3}{@{}p{\textwidth}@{}}{\textbf{Llama-3 (zero-shot), Hierarchical:} Mary is the daughter of Clotel, a slave who is sold to a slave trader. She is treated poorly by her mistress, but eventually gains her friendship. Mary is beautiful and intelligent, and she falls in love with George, a slave who is also her father's slave. She helps him escape from prison and they plan to run away together, but Mary is sold to a slave trader and taken to New Orleans. She is eventually reunited with George, and they get married.
 } \\
\midrule
\textbf{QA pairs:} &  &  \\
Who is Mary's mother? 	& Prediction: \textcolor{blue}{Clotel} &	 Answer: Clotel \\
Where is Mary sold after helping George escape? &	 Prediction: \textcolor{blue}{New Orleans} &	 Answer: New Orleans \\
Who rescues Mary from a similar fate as her mother? &	 Prediction: \textcolor{Red}{George} &	 Answer: Devenant \\
Who does Mary end up marrying? &	 Prediction: \textcolor{blue}{George} &	 Answer: George. \\
\midrule
\multicolumn{3}{@{}p{\textwidth}@{}}{\textbf{Llama-3, Lead:}  Mary is the daughter of Thomas Jefferson and a slave. She is a quadroon and is very beautiful. She is the sister of Clotel and Althesa. Mary is very intelligent and is able to read and write. She is also very knowledgeable about politics and is able to discuss the issues of the day with great skill. Mary is very determined and is able to overcome many obstacles in her life. She is also very kind and is able to help many people.} \\
\midrule
\textbf{QA pairs:} &  & \\
Who is Mary's mother? 	& Prediction: \textcolor{Red}{Thomas Jefferson} 	& Answer: Clotel \\
Where is Mary sold after helping George escape? 	& Prediction: \textcolor{Red}{slave} &	 Answer: New Orleans \\
Who rescues Mary from a similar fate as her mother? &	 Prediction: \textcolor{Red}{slave} & 	 Answer: Devenant \\
Who does Mary end up marrying? &	 Prediction: \textcolor{Red}{Thomas Jefferson} &	 Answer: George \\
\bottomrule
\end{tabular}
\caption{Examples of generated descriptions and questions with \textcolor{blue}{correct} and  \textcolor{red}{wrong} answers based on them.}
\label{tab:eval_examples2}
\end{table*}

\begin{table*}[ht]
\centering
\begin{tabular}{@{}p{0.5\textwidth} p{0.2\textwidth} p{0.25\textwidth}@{}}
\toprule
\multicolumn{3}{@{}l}{\textbf{Character:} Tita Bordereaux \quad \textbf{Book:} The Aspern Papers by Henry James} \\ 
\midrule
\multicolumn{3}{@{}p{\textwidth}@{}}{\textbf{Reference Description:} She is a tall, thin, middle-aged, American lady, and Juliana's niece. When the narrator broaches the subject of the papers to her, she timidly agrees to help him search for them. After her aunt dies, she cannot decide whether to respect Juliana's privacy, or give the narrator the papers. Towards the end, Tita puts forth the idea of marriage, but is rebuffed by the narrator, and she delights in telling him how she burnt the papers one by one. Seeing her in a new light, the narrator now claims that she is "plain, dingy, elderly," but not "hard or vindictive." In 1908, the Aspern Papers were revised, and Tita is called Tina.
} \\
\midrule
\multicolumn{3}{@{}p{\textwidth}@{}}{\textbf{Llama-3 + Coref (fine-tuned):} Tita Bordereaux is the niece of Juliana Bordereaux. She is a spinster and lives with her aunt in the old palace in Venice. She is shy and timid. She is the one who answers John Cumnor's letter, and she is the one who is responsible for showing the narrator around the palace. She is in love with the narrator and wants to marry him. She is very interested in flowers and gardening.
 } \\
\midrule
\textbf{Facts:} &  &  \\
Tita Bordereaux is the niece of Juliana Bordereaux.  & Prediction: \textcolor{blue}{True} &	 Type: Relationship   \\
Tita Bordereaux is a spinster. 	 	& Prediction: \textcolor{blue}{True} & Type: Other Fact \\
Tita Bordereaux lives with her aunt, Juliana Bordereaux, in the old palace in Venice.  	& Prediction: \textcolor{blue}{True}  &	 Type: Relationship\\
Tita Bordereaux is shy.  	& Prediction: \textcolor{blue}{True}  &	 Type: Personality \\
Tita Bordereaux is timid. 	& Prediction: \textcolor{blue}{True}   &	 Type: Personality \\ 
Tita Bordereaux answers John Cumnor's letter. 		& Prediction: \textcolor{red}{False}  & Type: Event  \\
Tita Bordereaux is responsible for showing the narrator around the palace. 	 	& Prediction: \textcolor{red}{False}   & Type: Event \\
Tita Bordereaux is in love with the narrator.  	& Prediction: \textcolor{red}{False}  &	 Type: Mental State \\
Tita Bordereaux wants to marry the narrator. 	 &	Prediction: \textcolor{red}{False}  & Type: Relationship \\
Tita Bordereaux is very interested in flowers. 	 &	Prediction: \textcolor{red}{False}  & Type: Other Fact \\
Tita Bordereaux is very interested in gardening.  &	 Prediction: \textcolor{red}{False}  &	 Type: Other Fact \\
Tita Bordereaux is very attached to her aunt, Juliana Bordereaux. 	 &	 Prediction: \textcolor{blue}{True} & Type: Relationship \\
Tita Bordereaux is very devoted. 		& Prediction: \textcolor{red}{False} & Type: Personality   \\
\bottomrule
\end{tabular}
\caption{Example of generated description with the corresponding extracted facts and their types. Facts are classified as \textcolor{blue}{supported} or \textcolor{red}{not} given the reference text.}
\label{tab:fact_eval_example1}

\end{table*}

\begin{table*}[ht]
\centering
\begin{tabular}{@{}p{0.5\textwidth} p{0.2\textwidth} p{0.25\textwidth}@{}}

\toprule
\multicolumn{3}{@{}l@{}}{\textbf{Character:} Gloucester \quad \textbf{Book:} King Lear by William Shakespeare} \\  
\midrule
\multicolumn{3}{@{}p{\textwidth}@{}}{\textbf{Reference Analysis:} Gloucester’s story runs parallel to Lear’s. Like Lear, Gloucester is introduced as a father who does not understand his children. He jokes about Edmund and calls him a “whoreson” (I.i.) when Edmund is standing right next to him. In his first soliloquy Edmund reveals how much he resents the way his father treats him. While the audience understands that Gloucester shouldn’t trust Edmund, Gloucester himself is blind to his son’s true motivations. Just as Lear falls for Goneril and Regan’s flattery, Gloucester falls for Edmund’s deception. Lear banishes Cordelia, the daughter who loves him, and Gloucester tries to execute Edgar, the son who loves him. Both Lear and Gloucester end up homeless, wandering on the beach near Dover. The close similarity between Gloucester’s story and Lear’s serves to underline that Lear’s fate is not exceptional. In the bleak universe of King Lear, it’s normal for old men to suffer at the hands of their own children and to end up with nothing.
The justness or unjustness of Gloucester’s fate remains unclear. Edmund, who deliberately sets out to destroy Gloucester, claims that he is acting in the name of natural justice: “Thou, Nature, art my goddess. To thy law / My services are bound” (I.ii.). Before he blinds Gloucester, Cornwall admits that it is unjust to harm him without a proper trial. Edgar argues that Gloucester deserves to lose his eyes for fathering an illegitimate son. Gloucester himself comes to believe that the world is unjust and cruel: “As flies to wanton boys are we to th’ gods / They kill us for their sport” (IV.i.). [truncated]
} \\
\midrule
\multicolumn{3}{@{}p{\textwidth}@{}}{\textbf{LongT5 + Coref (fine-tuned):} The character of Gloucester is a significant figure in Act I of "The Tragedy of King Lear". He is the Earl of Gloucester, a nobleman who is loyal to King Lear and has a deep sense of duty and responsibility. Gloucester is a complex character with a rich inner life, and his interactions with other characters reveal his thoughts, feelings, and motivations. Gloucester's relationship with Lear is one of loyalty and respect. He is deeply devoted to the king and is willing to do whatever it takes to support him, even if it means going against his own interests. This is evident when Gloucester agrees to help Lear by providing him with a map of the kingdom and offering to serve as his guide. Gloucester's loyalty to Lear is unwavering, and he is willing to risk his own life to protect the king. Gloucester's relationship with his illegitimate son, Edmund, is also significant. Edmund is a manipulative and deceitful character who seeks to overthrow his legitimate brother, Edgar, and claim the title of Earl of Gloucester for himself. Gloucester is unaware of Edmund's true nature and is initially fooled by his son's charm and wit. However, as the play progresses, Gloucester becomes increasingly suspicious of Edmund's motives and eventually discovers his treachery. Gloucester's character is also marked by his sense of duty and responsibility. He is deeply concerned about the welfare of his legitimate son, Edgar, and is determined to protect him from Edmund's machinations. Gloucester's sense of highlighting the king's flaws and weaknesses. Gloucester's loyalty and sense of duty are in stark contrast to Lear's foolishness and lack of judgment. Gloucester's character also serves as a reminder of the consequences of Edmund's treachery, and his fate serves as a warning to Lear and other characters about the dangers of unchecked ambition and deceit.
} \\
\midrule
\textbf{QA pairs:} &  &  \\
Who does Gloucester joke about and call a "whoreson"? 	& Prediction: \textcolor{blue}{Edmund} &	 Answer: Edmund \\
What does Gloucester realize about his son Edmund in his first soliloquy? 	 & Prediction: \textcolor{red}{manipulative and deceitful} 	 & Answer: Resents treatment  \\
Who does Gloucester fall for a deception similar to Lear falling for flattery? 	 & Prediction: \textcolor{blue}{Edmund} 	& Answer: Edmund \\
According to Edmund, who is his goddess? 	& Prediction: \textcolor{red}{Lear} &	 Answer: Nature \\
Who claims that Gloucester deserves to lose his eyes? 	& Prediction: \textcolor{red}{Edmund} 	 & Answer: Edgar \\
What does Gloucester come to believe about the world? 	& Prediction: \textcolor{red}{confused and uncertain}  &	 Answer: Unjust and cruel \\
What does Gloucester's blinding scene suggest about the act? &	 Prediction: \textcolor{red}{dangers} 	& Answer: Meaningless and random. \\
\bottomrule
\end{tabular}
\caption{Examples of generated analysis and questions with \textcolor{blue}{correct} and  \textcolor{red}{wrong} answers based on them.}
\label{tab:eval_examples4}
\end{table*}

\begin{table*}[ht]
\centering
\begin{tabular}{@{}p{0.45\textwidth} p{0.2\textwidth} p{0.3\textwidth}@{}}
\toprule
\multicolumn{3}{@{}l@{}}{\textbf{Character:} Jacomo \quad \textbf{Book:} The Jew of Malta by Christopher Marlowe} \\ 
\midrule
\multicolumn{3}{@{}p{\textwidth}@{}}{\textbf{Reference Analysis:} Jacomo is the friar who converts Barabas's daughter Abigail into a nun. Twice. And that's his high point: later, he and Friar Bernadine both get caught up in one of Barabas's murder plots after they (really awkwardly and incompetently) confront him about his involvement in the recent deaths in Malta.
And that's not to mention his greed. Jacomo and Bernadine are from rival orders, and when Barabas pretends to want to convert they resort to really petty squabbling over whose order will get to receive him and, more importantly, his money. Jacomo is ultimately a pretty sleazy guy. He all but licks his lips over Barabas's wealth. And then, when Barabas tricks him into believing that he's murdered Bernadine, his first thought is how to escape punishment. He even claims that, since he's a "sacred person," Barabas can't touch him.
In the end, Jacomo's brilliant plan of, um, running away doesn't work. Barabas successfully frames him for Bernadine's murder before the Maltese court and Ithamore later attends his execution.
Even though Jacomo isn't actually responsible Bernadine's death, you're not exactly crying when you find out he's been executed.
} \\
\midrule
\multicolumn{3}{@{}p{\textwidth}@{}}{\textbf{Llama-3 (fine-tuned):} Jacomo is a servant of the Governor. He is the one who tells Ferneze that the Jew is dead. He also tells Fernez that the Governor's son, Lodowick, is in the Jew's house. He then tells the Governor that the Turks are coming and the Governor tells him to warn the Knights. Jacomo is the only servant mentioned in the play. He has no lines and is only mentioned in passing. He does not appear on stage. He seems to be just a minor servant who is only there to carry out the orders of the high-ranking officials. He may be a representative of the common people, who are not really represented in the rest of the play. It is worth noting that Jacomo's name is Italian. This may be because the play is set in Malta and Italy is a neighboring country. It could also be because Marlowes' play is a comedy and he wanted to add a bit of Italian flavor to the play, even though it is set on the island of Malta and not in Italy. Marlow's use of Italian names and phrases in his play may be an attempt to make the play seem more exotic and exciting. It may also be a way for Marlow to show that the play's setting is near Italy and to add to the sense of place and setting that he is trying to create in the audience. The use of an Italian name for a servant may also add to Marlow’s attempt to show the multicultural nature of the society in which the play takes place. It is worth nothing that the other servants in the Governor’s household are called by English names. This could be a sign that Marlow is trying show that there are different cultures in the society of the time. It also shows that Marlov is trying use the play to comment on the society that he lives in. He uses the play as a way to comment of the multicultural society of his time. He shows that there is diversity in the world around him and he also shows the negative effects of this diversity. It could also mean that Marlo was trying to show how the society was changing. The play is written in the 16th century and the society at that time was changing and becoming more multicultural. Marlov may have been trying to comment in his own way on the changes that were happening in the time in which he lived. He could have been using the play in order to show his own views on the multiculturalism of his society} \\
\midrule
\textbf{QA pairs:} &  &  \\
Who converts Abigail into a nun twice? &	 Prediction: \textcolor{blue}{Jacomo} &	 Answer: Jacomo \\
What is Jacomo's reaction when he believes Barabas has murdered Bernadine? &	 Prediction: \textcolor{Red}{dead} &	 Answer: Escape punishment \\
Who successfully frames Jacomo for Bernadine's murder? 	& Prediction: \textcolor{Red}{Ferneze} 	& Answer: Barabas \\
How does Jacomo meet his fate? &	 Prediction: \textcolor{Red}{Turks} &	 Answer: Execution \\
\bottomrule
\end{tabular}
\caption{Examples of generated analysis and questions with \textcolor{blue}{correct} and  \textcolor{red}{wrong} answers based on them.}
\label{tab:eval_examples3}
\end{table*}

\begin{table*}[ht]
\centering
\begin{tabular}{@{}p{0.5\textwidth} p{0.2\textwidth} p{0.25\textwidth}@{}}
\toprule
\multicolumn{3}{@{}l@{}}{\textbf{Character:} Gloucester \quad \textbf{Book:} King Lear by William Shakespeare} \\ 
\midrule
\multicolumn{3}{@{}p{\textwidth}@{}}{\textbf{Llama + Coref (fine-tuned):} Gloucester is the most important character in the play. He is the one who is the only one who can be trusted to be honest and fair. He has a good sense of humor, and he has shrewd judgment. He knows what is right and wrong, and is able to make the right decision. He also has wit and wisdom, which is a great asset to the play, and helps to make it possible for the audience to get a sense of what is going on. Gloucester is incredibly well-known in the plays, but his character is not very well known. He's a rogue, and his actions are not particularly well-researched. He seems to be a bit of a joke, and it's not until we meet him that we see his true nature. He doesn't seem to be very much of racial or socially conservative, and yet he is genuinely a character. He does have a lot of good qualities, but he's also a little too nave for the play's purpose. He tries to be good and honest, but is unable to make up his mind. He can't trust his own judgment, and instead he tries a pretty witty approach to the text. He tells us that he doesn''t know what's right and what''s wrong, but that if he does, he will be honest. He always tries his best to tell us what'll be right and honest. Gloucester's actions are often a result of his own pranks, but they're also based on his own inexperience. He goes to Gloucester' s party, and then he takes a chance on him. In the end, despite his prickly attitude, a few things are afoot: he believes he can''ve done something wrong, if only he had been honest. It's hard to imagine a man who would have been honest and honest in the first place, but Gloucester isn't a bad guy. He takes pity for his sister's misfortune, and wants to be fair to her. He wants to make sure that she's treated fairly, and that a fair deal is made.
} \\
\midrule
Gloucester is the most important character in the play. 	& Prediction: \textcolor{red}{False} &	 Type: Role \\
Gloucester is the only character who can be trusted to be honest and fair. 	 & Prediction: \textcolor{red}{False} 	& Type: Role \\
Gloucester has a good sense of humor. &	 Prediction: \textcolor{blue}{True} &	 Type: Personality \\
Gloucester has shrewd judgment. &	 Prediction: \textcolor{red}{False} 	& Type: Personality\\
Gloucester knows what is right and wrong. &	 Prediction: \textcolor{red}{False} &	 Type: Mental State\\
Gloucester is able to make the right decision. &	 Prediction: \textcolor{red}{False} 	& Type: Mental State\\
Gloucester has wit. 	& Prediction: \textcolor{red}{False} &	 Type: Personality\\
Gloucester has wisdom. &	 Prediction: \textcolor{red}{False} 	& Type: Personality\\
Gloucester's wit and wisdom are great assets to the play. &	 Prediction: \textcolor{red}{False} &	 Type: Personality\\
Gloucester's wit and wisdom help the audience to understand what is going on in the play. &	 Prediction: \textcolor{red}{False} &	 Type: Other Fact\\
Gloucester is incredibly well-known in the plays. 	& Prediction: \textcolor{red}{False} &	 Type: Other Fact\\
Gloucester's character is not very well known. &	 Prediction: \textcolor{red}{False} 	& Type: Other Fact\\
Gloucester is a rogue character in the play. &	 Prediction: \textcolor{red}{False} 	 & Type: Role\\
Gloucester's actions are not particularly well-researched. &	 Prediction: \textcolor{red}{False} 	 & Type: Other Fact\\
Gloucester seems to be perceived as a bit of a joke. 	& Prediction: \textcolor{blue}{True} 	& Type: Personality\\
Gloucester's true nature is revealed upon meeting him. &	 Prediction: \textcolor{red}{False} &	 Type: Personality\\
Gloucester does not seem to be racially conservative. 	& Prediction: \textcolor{red}{False} 	& Type: Personality\\
Gloucester does not seem to be socially conservative. &	 Prediction: \textcolor{blue}{True} 	& Type: Personality\\
Gloucester is a genuine character. 	& Prediction: \textcolor{red}{False} 	& Type: Other Fact\\

[Facts Truncated] & & \\

\bottomrule
\end{tabular}
\caption{Example of generated analysis with the corresponding extracted facts and their types. Facts are classified as \textcolor{blue}{supported} or \textcolor{red}{not} given the reference text. The gold-standard of this example can be found in Table~\ref{tab:eval_examples4}.}
\label{tab:fact_eval_example2}
\end{table*}

\begin{table*}[ht]

\centering
\begin{tabular}{@{}p{0.98\textwidth}@{}}
\toprule
\textbf{Characters:}  Nameless Captain, Leggatt, Captain Archbold, The Chief Mate and The Steward \\  \textbf{Book:} Secret Sharer by Joseph Conrad \\
\midrule
\textbf{Reference:} \\
\textbf{Nameless Captain:} The narrator of the story, this is a young man who only a fortnight before the story begins receives his first command of a ship. Because he has been on the ship only two weeks, at the beginning of the novel he feels like he is not only a "stranger to the ship" but also a "stranger to himself."  [truncated] \\
\textbf{Leggatt:} An escaped criminal, it is his presence on the nameless ship that brings crisis to the captain and his ship. Originally, a first mate on the nearby Sephora, during a storm Leggatt, in a just rage, murdered an inferior, but in the process, saved the ship. Unappreciated, he is locked up for weeks before he escapes the ship and swims to nearby ship on which the captain resides. Like the captain, he is a young man and went to the same boy's prep school as his "secret self." [truncated] \\
\textbf{Captain Archbold:} The captain of the Sephora, many critics believe that he is the true villain of The Secret Sharer. With red whiskers, he is nervous and scared of all that is on his ship, including the first mate, Leggatt, and even his wife. Harsh and unwilling to compromise or admit that he is wrong regarding his stance towards Leggatt, he is a foil to both the captain and Leggatt as someone who is strictly law obeying. During his conversation with the captain, he gives off an air of fussy distraction, and in his most authoritative act sticks out his tongue to imitate the death mask of Leggatt's victim. [truncated] \\
\textbf{The Chief Mate:} Throughout the narrative, the captain refers to this character as "terrible whiskers," "frightful whiskers," and "terrific whiskers." The captain considers every occurrence on the ship, trying to figure out they why and how of things. Most identifiable, he is concerned with why a scorpion chose his cabin to drown in an inkwell.
 \\
\textbf{The Steward:} His main function is plot, as the character who comes in closest contact with the captain, he is most likely to discover the secret of Leggatt. Primarily because of this function, the captain becomes rude and surly towards him throughout the course of the narrative.
\\
\midrule
\textbf{Llama-3 (zero-shot), Hierarchical:} \\
\textbf{Nameless Captain:} The protagonist of the story, a young captain who has just taken command of a ship. He is reflective and introspective, and is still getting used to his new role. He is haunted by the presence of his double, Leggatt, who appears on his ship. \\
\textbf{Leggatt:} A young man who has been a mate on the ship Sephora. He is a strong swimmer and has a mysterious past, having killed a man on his previous ship. He is calm and resolute, and has a sense of self-possession. He is the captain's double, and his presence on the ship creates a sense of unease and tension. \\
\textbf{Captain Archbold:} The captain of the Sephora, Leggatt's previous ship. He is not physically present in the story, but is mentioned as being a sickly man who is affected by the events on his ship. He is a seasoned sailor with many years of experience, but is also shown to be somewhat dense and unimaginative. \\
\textbf{The Chief Mate:} An older man with a thick beard and a simple, straightforward way of thinking. He is a bit of a worrier and is prone to making pronouncements. He is often at odds with the captain, who he sees as inexperienced and reckless. \\
\textbf{The Steward:} A minor character who is responsible for serving the captain and officers. He is not mentioned much in the story, but plays an important role in facilitating the secret conversations between the captain and Leggatt.
\\

\bottomrule
\end{tabular}
\caption{Joint character description, gold-standard (top) and system output (bottom).}
\label{tab:joint_example}
\end{table*}

\clearpage
\newpage

\end{document}